\documentclass[11pt]{article}
\usepackage[utf8]{inputenc}
\usepackage[T1]{fontenc}
\usepackage[margin=1in]{geometry}
\usepackage{amsmath,amssymb,bm}
\usepackage{graphicx}
\usepackage{booktabs}
\usepackage{microtype}
\usepackage[hidelinks,hypertexnames=false]{hyperref}
\usepackage{caption}
\DeclareMathOperator{\arccosh}{arccosh}
\newcommand{\dFR}{d_{\mathrm{FR}}}

\title{Fisher--Rao Geodesic Attention: Spherical and Hyperbolic Attention from Information Geometry}
\author{Eduardo Qui\~nelen Rojas \qquad Miguel Contreras Torrejon\\[4pt]
\small Independent researchers, Santiago, Chile\\
\small Code: \url{https://github.com/quinelen/fisher-rao-attention}\\
\small Concept DOI: \href{https://doi.org/10.5281/zenodo.23424909}{10.5281/zenodo.23424909}}
\date{July 2026 (v1); November 2026 (this version)}

\begin{document}
\maketitle

\begin{abstract}
Standard attention scores tokens by inner products in a flat space. We propose attention in which queries and keys are probability distributions and the score is their exact Fisher--Rao geodesic distance. Once the distributional family is chosen, invariance under sufficient statistics fixes the metric up to scale: categorical tokens yield a spherical kernel ($K=+1/4$), isotropic Gaussian tokens a hyperbolic one ($\mathbb{H}^{d+1}$), and Euclidean distance attention is the flat Gaussian location family. The isotropic Gaussian kernel is hyperbolic distance attention in half-space coordinates; its exact distance needs one matrix multiplication plus an elementwise $\arccosh$ ($2.3$--$2.8\times$ a dot-product attention module on CPU). On Dyck-2 with matched parameters and a fixed training budget, closing-bracket accuracy beyond the training length ranks dot-product $85.8\% <$ spherical $86.6\% <$ hyperbolic $88.4\%$; the margin is the same within the training length, so it reflects better learning of nested structure rather than a separate extrapolation effect. The contrast holds over 20 paired seeds ($+2.10$ pp, 95\% CI $[1.74, 2.46]$), is larger in one 19M-parameter configuration ($+4.7$ pp, 3 seeds) and persists with rotary positions ($+1.82$ pp, $n{=}20$, pre-registered). With Gaussian tokens held fixed, the Fisher--Rao kernel beats Wasserstein--Bures ($+1.52$ pp) and KL ($+1.11$ pp) kernels; KL shares its local geometry but grows quadratically where the geodesic grows logarithmically, and a variant without per-token scales keeps most of the gain in an unpaired comparison, so the evidence points to the geodesic kernel rather than to Gaussian tokens per se, without yet separating its curvature from its slow growth. On character-level code and technical English prose, with learned absolute positions whose rows beyond the training length are never trained, a norm-capped variant degrades less than dot-product attention beyond the training length (pre-registered; $-0.084$ and $-0.058$ bpc), but not on Spanish clinical text ($p = 0.45$); these small CPU models remain near character-bigram quality, so this is not evidence about natural-language extrapolation at scale. Pre-registered probes of the learned scales find that second-layer query scales narrow and key scales widen with nesting depth in every head and seed, a strongly decodable code whose removal from the query scales at test time costs $0.15$ pp beyond the training length and $0.41$ pp within it; its cause remains open. A reimplementation of curvature-adaptive routing (CAT) shows no significant difference ($+0.30$ pp, 95\% CI $[-0.91, 1.51]$). Negative results are reported alongside the positive ones.
\end{abstract}


\section{Introduction}

Scaled dot-product attention measures similarity in a flat vector space whose metric depends entirely on the learned parameterization: its similarity has no intrinsic metric meaning. A long line of work has argued that hierarchical structure --- trees, nested grammars, ontologies --- embeds with low distortion only in negatively curved spaces \cite{nickel2017,gulcehre2019,gu2019}, and hyperbolic networks have shown gains on hierarchy-heavy tasks. Their adoption, however, has been limited by two frictions: the geometry is imposed \emph{a priori} (the designer must know the data is hierarchical), and hyperbolic operations have historically carried significant computational overhead.

This paper addresses both frictions inside a single formalism. Our starting point is a different answer to what a query and a key are: not points compared by an inner product, but members of a statistical family compared by the Fisher--Rao distance. Tokens as distributions are not new \cite{vilnis2015,erick2023stochastic,fan2022stosa}; what follows from requiring invariance under sufficient statistics is that the family fixes the metric up to scale, so that it is no longer a separate choice. Two consequences are worked out below: the three constant-curvature geometries become cases of one construction (Section~\ref{sec:members}), and, in the isotropic Gaussian family, each token acquires an explicit scale parameter, which point tokens do not have and whose learned meaning is an empirical question (Section~\ref{sec:sigma}).

We are careful to separate what is novel here from what re-instantiates known ideas (see Section~\ref{sec:related} for detailed positioning; adaptive mixed-curvature attention in particular is prior work \cite{lin2025cat,cho2023curve,he2025helm}, not a contribution of this paper). Our contributions:

\begin{enumerate}
\item \textbf{An information-geometric unification of attention geometries.} Mapping queries and keys to probability distributions and equipping the family with the Fisher--Rao metric reduces geometry selection to a single principled design choice --- the distributional family --- after which the metric is fixed up to scale by invariance under sufficient statistics (Chentsov's theorem \cite{chentsov1972} and its extension to parametrized measure models \cite{ay2015}). Categorical families yield spherical geometry, isotropic Gaussian families yield genuinely hyperbolic geometry, and Euclidean distance attention is the flat Gaussian location family at fixed $\sigma$, a horosphere of $\mathbb{H}^{d+1}$ (not a point-mass limit, which is singular --- Fisher information diverges as $\sigma \to 0$). We are explicit that this shifts rather than eliminates design freedom: the family choice, and the dimension-independent curvature convention of Eq.~(1$'$), remain modeling decisions. The contrast with prior mixed-curvature transformers, which select model spaces (Poincar\'e/Lorentz/hypersphere) with fixed or learnable curvature \cite{lin2025cat,cho2023curve,he2025helm}, is that here the sign of the curvature is a \emph{consequence} of a statistical modeling choice; its magnitude is a units convention absorbed by the learned temperature, as it also is for model-space hyperbolic attention.
\item \textbf{An exact isotropic-Gaussian hyperbolic kernel computed with one matrix multiplication.} The closed-form Fisher--Rao geodesic between isotropic Gaussian tokens (Eq.~\ref{eq:dist}) reduces its pairwise computation to one matrix multiplication plus $O(L^2)$ scalar operations (Section~\ref{sec:iso}): $2.3$--$2.8\times$ the cost of a standard attention module across three machines ($\approx 1.4\times$ in end-to-end training time), versus $28$--$110\times$ for the naive per-coordinate product of half-planes (Appendix~\ref{app:cost}). The reduction is the Minkowski bilinear form of the hyperboloid, which Lorentz-model attention also uses \cite{chen2022fully} (Section~\ref{sec:related}).
\item \textbf{Controlled evidence on where the gain comes from.} On Dyck-2 with parameter-matched models and a fixed training budget, closing accuracy ranks dot-product $<$ spherical $<$ hyperbolic, with similar margins within and beyond the training length. The hyperbolic-isotropic mechanism is ahead of every other arm --- significantly against the Euclidean, spherical and Ricci-approximation arms (only the Euclidean contrast survives Bonferroni correction) and directionally against the larger anisotropic variant ($p = 0.063$) --- and its contrast with Euclidean attention holds over 20 paired seeds and is larger in one $46\times$ (19M-parameter) configuration (Section~\ref{sec:experiments}, Section~\ref{sec:extended}a--b). Ablations that hold the Gaussian token representation fixed show that the gain is not carried by Gaussian tokens per se: Fisher--Rao beats both a Wasserstein--Bures kernel and a KL (Bregman) kernel that shares its local information geometry (Section~\ref{sec:wasserstein}). What these ablations isolate is the geodesic kernel and its slow (logarithmic) growth; separating curvature from that growth needs controls we have not run (Section~\ref{sec:limitations}). With relative (rotary) positions in both arms, the advantage persists (Section~\ref{sec:extended}f).
\item \textbf{A threshold-gated parameter/compute efficiency result.} Above a head-width threshold ($d_h \ge 16$), the hyperbolic-isotropic kernel reaches accuracy comparable to (not significantly different from) Euclidean attention at double the dimension using $\approx 3.8$--$3.9\times$ fewer parameters in a short-training sweep and, at the largest scale tested, less real wall-clock training time; below the threshold the effect reverses (Section~\ref{sec:efficiency}). This is the paper's only positive resource-efficiency claim against Euclidean attention, after three other efficiency/advantage hypotheses were tested and refuted in follow-up work (graph embedding, Section~\ref{sec:incumbents}; few-shot classification, whose script is in the project repository; uncertainty calibration, run in a separate codebase that is not included --- the latter two out of scope for the attention mechanism proper).
\item \textbf{A negative result on the necessity of geometry gating.} Re-instantiating the adaptive mixed-curvature idea \cite{lin2025cat} in our information-geometric setting, we find (via a pre-specified stress test, Section~\ref{sec:stress}) that the pure hyperbolic kernel matches or beats the gated mixture even on structurally heterogeneous data --- suggesting negative curvature functions as a superset default rather than requiring explicit routing. A pre-registered comparison against a reimplementation of the central mechanism of the incumbent curvature-adaptive transformer (CAT) on Dyck-2 --- to our knowledge the first evaluation of CAT's per-token routing on a sequence task --- finds no significant difference from the pure kernel (no equivalence margin was pre-specified), which is simpler and faster (Section~\ref{sec:cat}). The gate remains useful as a diagnostic probe of data structure (Section~\ref{sec:gate}).
\item \textbf{Evidence on what the learned scale encodes.} Pre-registered probes show that the per-token scale encodes nesting depth in the second layer, with query scales narrowing and key scales widening, on Dyck-2, on a four-type variant with the same depth sequences and on a deeper Dyck-2 distribution (Section~\ref{sec:sigma}). The code is strongly decodable, but removing it from the query scales at test time costs 0.15 pp beyond the training length and 0.41 pp within it; it was not detected on program source at this scale, and its cause remains open.
\end{enumerate}

\section{Background: One Theorem, Three Curvatures}\label{sec:background}

\subsection{Attention heads as maps to statistical manifolds}

Let $x \in \mathbb{R}^{d_{\mathrm{model}}}$ be a token embedding. Each head projects tokens into a parametric family of probability distributions $p(z\mid\theta)$ rather than into raw vectors: the head outputs the parameter $\theta$ of one member of the family, and the sample variable $z$ is auxiliary and is never drawn. The Fisher information metric
\begin{equation*}
g_{\mu\nu}(\theta) = \mathbb{E}_{z\sim p(\cdot\mid\theta)}\!\left[\partial_\mu \log p(z\mid\theta)\, \partial_\nu \log p(z\mid\theta)\right]
\end{equation*}
equips the family with a Riemannian structure \cite{amari2016}. By \textbf{Chentsov's theorem} for finite sample spaces \cite{chentsov1972} and its extension to parametrized measure models \cite{ay2015}, the Fisher metric is, up to a universal constant, the only family of Riemannian metrics on statistical models invariant under sufficient statistics: \emph{given the distributional family}, the geometry of the head is fixed up to scale. Since scale is absorbed by the learned temperature (Section~\ref{sec:members}), what the family determines is the sign of the curvature and the chart in which tokens are parameterized. The family itself is the designer's choice --- the framework's contribution is to concentrate all geometric freedom into that one statistically interpretable decision.

\subsection{The three constant-curvature members}\label{sec:members}

\begin{figure}[t]
\centering
\includegraphics[width=\textwidth]{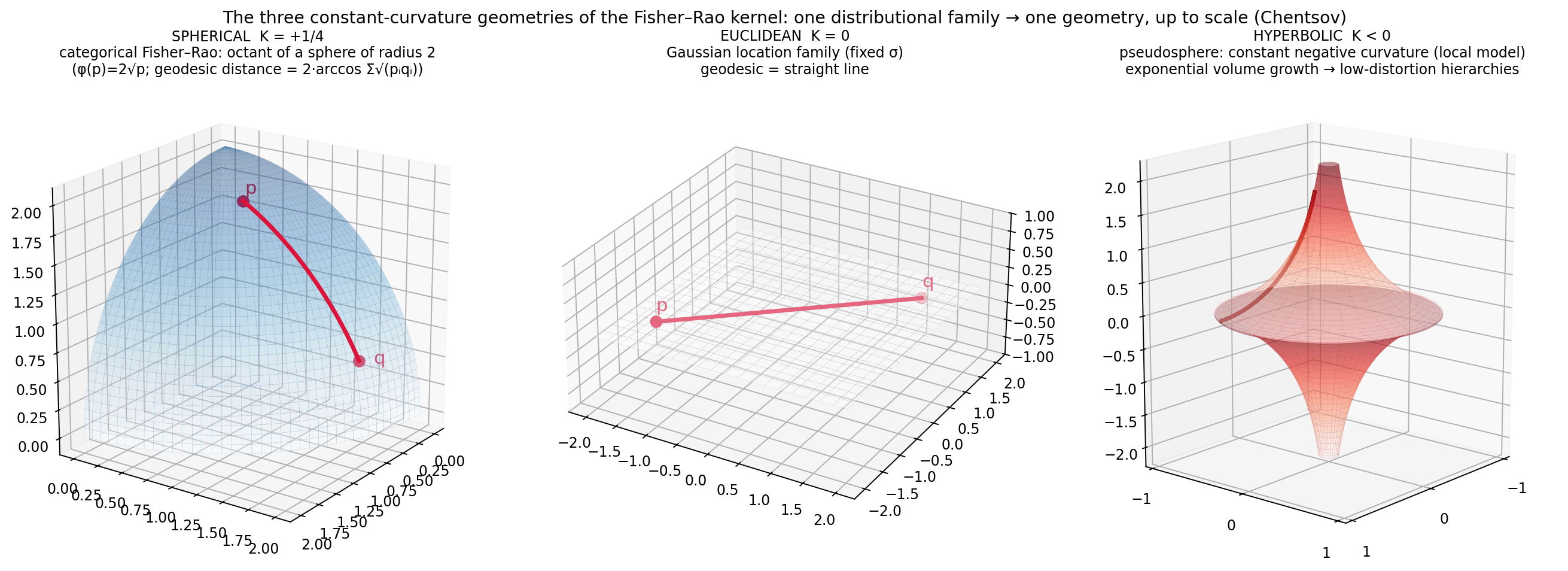}
\caption{The three constant-curvature geometries of the family: the categorical Fisher--Rao octant (left, with an exact great-circle geodesic between two distributions), the flat member, the Gaussian location family at fixed $\sigma$ (center), and constant negative curvature (right, a pseudosphere, which is only a local model of the hyperbolic plane; Figure~\ref{fig:halfspace} shows the global half-space model). An interactive 3D version of the curved scenes (this pseudosphere and Figures~\ref{fig:realtokens} and~\ref{fig:halfspace}) is included in the repository (\texttt{verificacion\_claude/}\allowbreak\texttt{geometrias\_interactivas.html}).}
\label{fig:geometries}
\end{figure}

Table~\ref{tab:members} summarizes the three members; Figure~\ref{fig:geometries} renders their manifolds.

\begin{table}[h]
\centering\small
\begin{tabular}{p{3.6cm}p{3.6cm}cp{4.6cm}}
\toprule
Distributional family & Manifold & Curvature & Closed-form geodesic distance \\
\midrule
Categorical $p \in \Delta^{d-1}$ & octant of sphere of radius 2 (via $\phi(p) = 2\sqrt{p}$) & $K = +1/4$ & $d = 2\arccos\big(\sum_i \sqrt{q_i k_i}\big)$ \\
Isotropic Gaussian $\mathcal{N}(\mu, \sigma^2 I_d)$ & hyperbolic space $\mathbb{H}^{d+1}$ & $K = -1/2$ & Eq.~\eqref{eq:dist} \\
Gaussian location family $\mathcal{N}(\mu, \sigma_0^2 I)$, $\sigma_0$ fixed & $\mathbb{R}^d$ (a horosphere of $\mathbb{H}^{d+1}$) & $0$ & $\lVert\Delta\mu\rVert / \sigma_0$ (Euclidean distance attention; the $\sigma_0 \to 0$ point-mass limit is singular) \\
\bottomrule
\end{tabular}
\caption{The three constant-curvature members of the Fisher--Rao attention family.}
\label{tab:members}
\end{table}

\begin{figure}[t]
\centering
\includegraphics[width=0.62\textwidth]{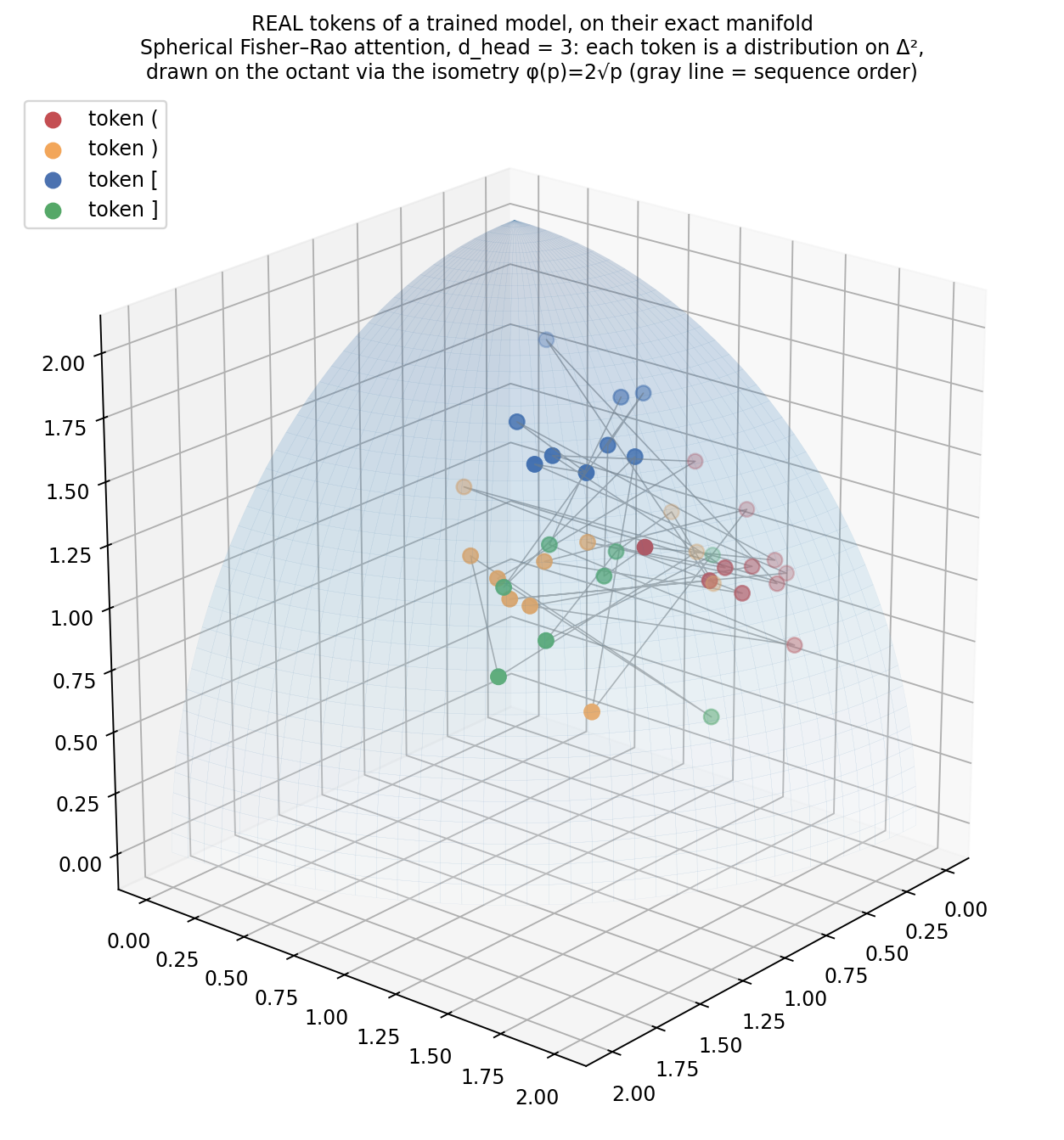}
\caption{Real token distributions from a trained spherical-attention model ($d_h = 3$) rendered on their exact manifold via $\phi(p) = 2\sqrt p$ --- no dimensionality reduction involved; the gray line traces sequence order.}
\label{fig:realtokens}
\end{figure}

The categorical (spherical) case: the map $\phi(p) = 2\sqrt{p}$ is an isometry from $(\Delta^{d-1}, g_F)$ onto the positive octant of the radius-2 sphere; the geodesic distance is the Bhattacharyya angle, bounded in $[0, \pi]$. Figure~\ref{fig:realtokens} shows token distributions from a trained model rendered exactly on this manifold.

\begin{figure}[t]
\centering
\includegraphics[width=0.72\textwidth]{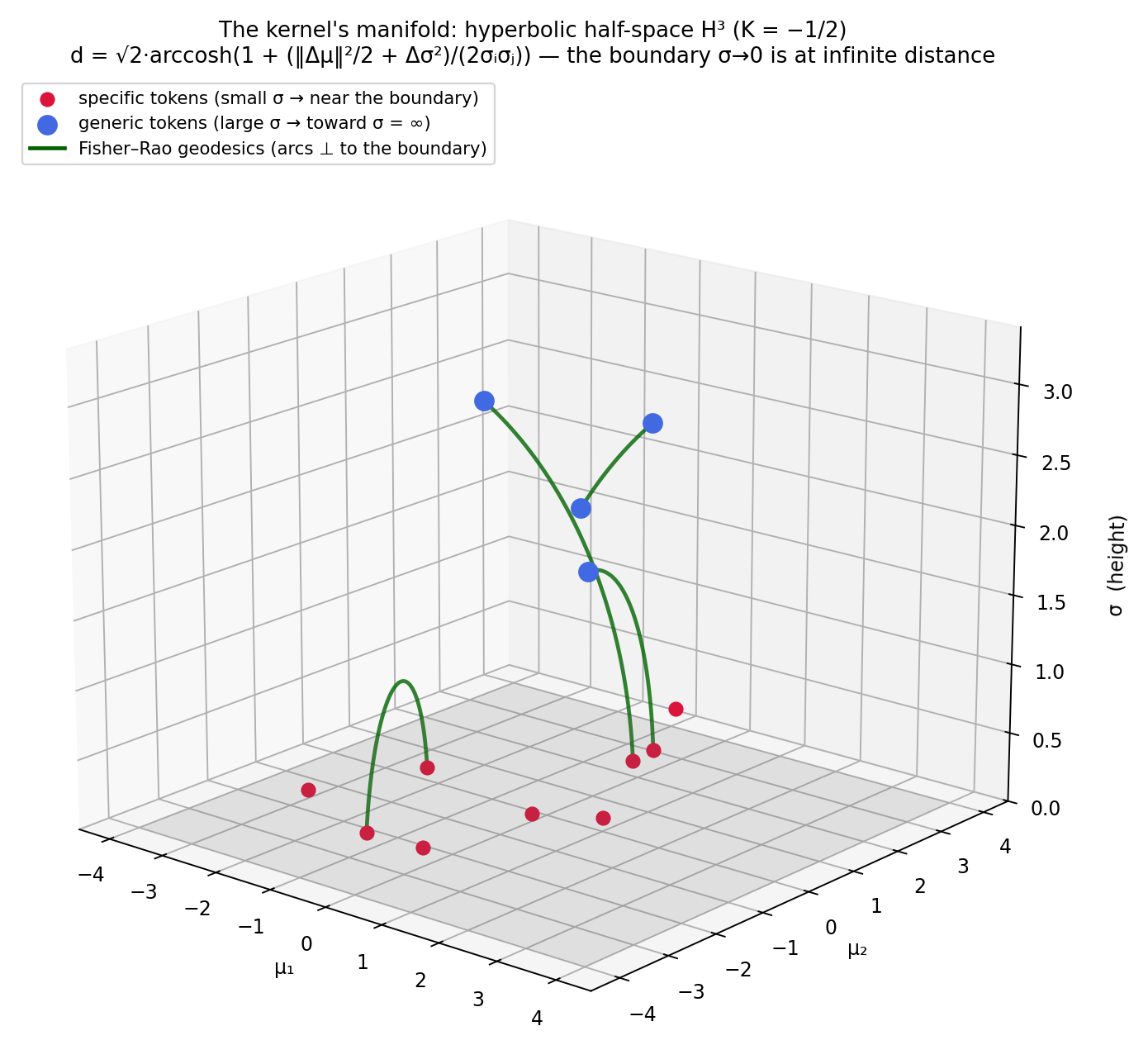}
\caption{The hyperbolic half-space manifold of the isotropic Gaussian kernel, with exact Fisher--Rao geodesics (arcs orthogonal to the boundary) and the geometric reading of $\sigma$ as height: specific tokens (small $\sigma$) near the boundary, generic tokens (large $\sigma$) toward the ideal point $\sigma = \infty$ (trained models use this axis differently for queries and keys; Section~\ref{sec:sigma}).}
\label{fig:halfspace}
\end{figure}

The isotropic Gaussian (hyperbolic) case: for $\mathcal{N}(\mu, \sigma^2 I_d)$ with $\mu \in \mathbb{R}^d$, $\sigma > 0$, the exact Fisher line element carries a dimension factor on the scale coordinate \cite{atkinson1981},
\begin{equation}\tag{1}\label{eq:metric1}
ds^2 = \frac{\lVert d\mu \rVert^2 + 2d\,d\sigma^2}{\sigma^2},
\end{equation}
which after rescaling $\sigma' = \sqrt{2d}\,\sigma$ becomes $2d$ times the standard upper half-space metric, i.e.\ hyperbolic space of curvature $K = -1/(2d)$.

\paragraph{Convention (dimension-independent curvature) --- and why it is a choice of units, not a departure from the theory.} Our kernel fixes the scale factor at its $d = 1$ value, using
\begin{equation}\tag{1$'$}\label{eq:metric1p}
ds^2 = \frac{\lVert d\mu \rVert^2 + 2\,d\sigma^2}{\sigma^2},
\end{equation}
which is hyperbolic with $K = -1/2$ \emph{independently of the head width} $d_h$, keeping the manifold constant across configurations (under the literal factor $2d$, curvature would flatten as $K = -1/(2d_h)$ with growing width, confounding cross-scale comparisons). Crucially, (1$'$) is \textbf{not} a different geometry from the genuine multivariate Fisher--Rao metric (1): pulling (1$'$) back through the linear reparameterization $\tilde\mu = \mu/\sqrt{d}$ gives exactly $\frac{1}{d}\times$ metric (1), a global homothety. Consequently $d^2_{(1)}(\mu, \sigma) = d \cdot d^2_{(1')}(\mu/\sqrt{d}, \sigma)$ identically, and since the projection $W_\mu$ is learned (absorbing the $1/\sqrt d$ rescaling) and the temperature $\beta$ is learned (absorbing the global factor $d$), \emph{the attention function class defined by (1$'$) is identical to the one defined by the literal metric (1)}. The same argument covers the $1/d_h$ logit normalization in Section~\ref{sec:iso}: a fixed constant absorbed by $\beta$. The convention fixes units so that reported curvatures and temperatures are comparable across widths; it does not trade theoretical fidelity for engineering convenience. To be exact about what is and is not invariant: the \emph{magnitude} of $K$ is a units convention (any homothety rescales it), while the \emph{sign} of the curvature --- negative for the Gaussian family, positive for the categorical, zero for the location family --- is invariant under homothety and reparameterization, and it is the sign, not the magnitude, that every experimental claim in this paper rests on. All curvature values reported for the Gaussian case, and the distance below, refer to (1$'$). The exact geodesic distance for this metric is \cite{costa2015}
\begin{equation}\tag{2}\label{eq:dist}
\dFR\big(({\mu}_1, \sigma_1), ({\mu}_2, \sigma_2)\big) = \sqrt{2}\,\arccosh\!\left(1 + \frac{\tfrac{1}{2}\lVert \mu_1 - \mu_2 \rVert^2 + (\sigma_1 - \sigma_2)^2}{2\,\sigma_1 \sigma_2}\right).
\end{equation}

Sanity limits: for $\mu_1 = \mu_2$ it reduces to $\sqrt{2}\,\lvert \ln(\sigma_2/\sigma_1)\rvert$; for $\sigma_1 = \sigma_2 = \sigma$ it is
\[
2\sqrt{2}\,\mathrm{arcsinh}\big(\lVert\Delta\mu\rVert / (2\sqrt{2}\,\sigma)\big),
\]
a monotone function of the mean separation in noise units, which tends to $\lVert\Delta\mu\rVert/\sigma$ for small separations (the hyperbolic distance on a horosphere, against the flat distance of the location family); it diverges logarithmically as $\sigma \to 0$ --- near-deterministic tokens live near the hyperbolic boundary. The scalar $\sigma$ is the height in the half-space, and $\log\sigma$ is, up to the factor $\sqrt{2}$, a horospherical (Busemann) coordinate toward the ideal point $\sigma = \infty$. Tokens with small $\sigma$ approach the finite boundary $\mathbb{R}^d \times \{0\}$, where the distance between tokens at a fixed mean separation grows without bound; tokens with large $\sigma$ approach the ideal point, where that distance shrinks ($\approx \lVert\Delta\mu\rVert/\sigma$). Ordering by $\sigma$ is the horospherical analogue of the radial ordering that Poincar\'e embeddings use for generality \cite{nickel2017}, and Tifrea et al.\ read generality off Gaussian variance in the Fisher geometry of diagonal Gaussians, a product of hyperbolic half-planes \cite{tifrea2019} (Figure~\ref{fig:halfspace}). Because dilations $(\mu, \sigma) \mapsto (\lambda\mu, \lambda\sigma)$ are isometries, only ratios of scales, and of scale to mean separation, are meaningful. The reading describes the geometry, not what a trained model does with it: on nested brackets, trained query scales follow it (narrower with depth) while key scales invert it (wider with depth; Section~\ref{sec:sigma}).

\section{Mechanisms}\label{sec:mechanisms}

All variants share the standard multi-head skeleton (projections, causal mask, softmax over negative squared distance, value aggregation); only the score computation differs.

\subsection{Spherical Fisher--Rao attention (categorical, exact, one matmul)}\label{sec:spherical}

\begin{equation*}
q = \mathrm{softmax}(W_Q x), \quad k = \mathrm{softmax}(W_K x) \in \Delta^{d_h - 1}, \qquad s_{ij} = -\,\dFR^2(q_i, k_j) = -4\arccos^2\!\big(B_{ij}\big),
\end{equation*}
where $B = \sqrt{Q}\,\sqrt{K}^{\top}$ is a single matrix multiplication of element-wise square roots. Because $\dFR \in [0, \pi]$, logits are intrinsically bounded and the kernel trains stably without a temperature parameter (Section~\ref{sec:ablations}). Implementation note: $\arccos$ has a divergent gradient at $B \to 1$, although the squared distance used in the score does not ($\mathrm{d}\arccos^2 B/\mathrm{d}B \to -2$); we clamp $B \le 1 - 10^{-4}$ against rounding.

\subsection{Hyperbolic Gaussian attention, anisotropic (product of half-planes)}\label{sec:aniso}

Each head projects to $d_h$ univariate Gaussians per token ($\mu = W_\mu x$, $\sigma = \mathrm{softplus}(W_\sigma x) + \varepsilon$), with squared distance summed per coordinate over the product manifold $(\mathbb{H}^2)^{d_h}$. This is the direct construction but costs $O(L^2 d_h)$ element-wise $\arccosh$ operations ($28$--$110\times$ Euclidean wall-clock across three machines; Appendix~\ref{app:cost}) and, as tested, carries extra parameters (separate $\sigma$ projections). We include it as an arm of the benchmark; the isotropic variant below is $12$--$44\times$ cheaper, parameter-matched, and at least as accurate (Section~\ref{sec:stats}).

\subsection{Hyperbolic isotropic attention (ours; exact, one matmul plus an elementwise arccosh, parameter-matched)}\label{sec:iso}

Take $\sigma$ \textbf{scalar per token per head}, with separate projections for the query scale $\sigma_q$ and the key scale $\sigma_k$: the token is an isotropic Gaussian living on $\mathbb{H}^{d_h+1}$ (under convention (1$'$), $(\mu, \sigma)$ represents $\mathcal{N}(\sqrt{d_h}\,\mu, \sigma^2 I_{d_h})$ with distances scaled by $1/\sqrt{d_h}$; Section~\ref{sec:members}). The only pairwise vector quantity in Eq.~\eqref{eq:dist} is $\lVert \mu_i - \mu_j \rVert^2 = \lVert\mu_i\rVert^2 + \lVert\mu_j\rVert^2 - 2\,\mu_i^{\top}\mu_j$ --- one matrix multiplication. The $\arccosh$ then acts on an $[L \times L]$ matrix rather than $[L \times L \times d_h]$:
\begin{equation*}
s_{ij} = -\beta\, \frac{2}{d_h} \arccosh^2\!\left(1 + \frac{\tfrac{1}{2}\lVert\Delta\mu_{ij}\rVert^2 + \Delta\sigma_{ij}^2}{2 \sigma_i \sigma_j}\right).
\end{equation*}
Numerical safeguards: the $\arccosh$ argument is clamped to $\ge 1 + 10^{-7}$, and $\sigma \ge \varepsilon$ keeps tokens off the infinitely-distant boundary. On the clamp, a point worth making precisely: the gradient singularity of $\arccosh$ at coincident tokens afflicts the \emph{distance} $d$, not the \emph{squared} distance the scores actually use --- $\frac{\partial d^2}{\partial A} = 4\arccosh(A)/\sqrt{A^2-1} \to 4$ (finite) as $A \to 1^+$, so the score gradients are smooth on the diagonal and the clamp guards only against floating-point arguments dipping below 1; it affects only arguments within about $10^{-7}$ of 1, so its gradient bias is negligible. For length-extrapolation robustness, a third safeguard is recommended: a \textbf{$\mu$-norm ball cap} ($\lVert\mu\rVert \le c$, e.g.\ $c = 12$), which Section~\ref{sec:extended}d shows stabilizes far-extrapolation behavior on program source under out-of-distribution positional inputs --- the analogue of Poincar\'e ball clamping. Parameter cost matches standard attention (the $\sigma$ projections map $d_{\mathrm{model}} \to n_{\mathrm{heads}}$: 415k vs.\ 413k parameters in our configuration). Measured cost of the attention module on the original hardware: $2.3\times$ Euclidean ($\approx 1.4\times$ in end-to-end training time, Section~\ref{sec:cat}), $12\times$ faster than the anisotropic version ($2.3$--$2.8\times$ and $12$--$44\times$ across three machines; Appendix~\ref{app:cost}).

\subsection{Mixed-curvature attention with causal data-driven gating}\label{sec:mixed}

Each head computes all three scores --- $s^{\mathrm{euc}}$ (dot product), $s^{\mathrm{sph}}$ (Section~\ref{sec:spherical}), $s^{\mathrm{hyp}}$ (Section~\ref{sec:iso}, sharing the same $q k^{\top}$ matmul) --- and blends them per position:
\begin{equation*}
s_{ij} = \alpha^{\mathrm{euc}}_i\, s^{\mathrm{euc}}_{ij} + \alpha^{\mathrm{sph}}_i\, s^{\mathrm{sph}}_{ij} + \alpha^{\mathrm{hyp}}_i\, s^{\mathrm{hyp}}_{ij}, \qquad \alpha_i = \mathrm{softmax}\big(W_g\, \bar{x}_{\le i}\big) \in \Delta^2,
\end{equation*}
where $\bar{x}_{\le i}$ is the \textbf{cumulative (causal) mean} of the input up to position $i$ --- the gate observes only the past, preserving autoregressive validity. Per-head, per-geometry learnable temperatures calibrate the heterogeneous logit scales. The gate values $\alpha$ are also the mechanism's built-in interpretability probe: they report, per position, head, and layer, which geometry the model is \emph{choosing to think in}.

\section{Experiments}\label{sec:experiments}

\subsection{Protocol}\label{sec:protocol}

\textbf{Task.} Dyck-2 (balanced sequences over \texttt{()} and \texttt{[]}), autoregressive next-token training with strict causal masking. Models train on lengths 20--60 and are evaluated out-of-distribution (OUT) on lengths 80--120. The discriminative metric is \textbf{bracket-closing type accuracy}: at positions whose target is a closing bracket, whether the model prefers the correct closing type --- a direct probe of stack tracking; cross-entropy is reported as a calibration measure. Dyck languages are a standard probe of how attention tracks hierarchical structure \cite{yao2021,hahn2020,bhattamishra2020,ebrahimi2020}.

\textbf{Baseline, seeds and reporting.} The flat baseline (``Euclidean'') is scaled dot-product attention \cite{vaswani2017}, not the Euclidean-distance attention of the flat Gaussian location family (Table~\ref{tab:members}); the two differ by a key-norm term \cite{kim2021lipschitz}. Seeds vary initialization and batch order; the Dyck-2 training and evaluation sets are generated once, with fixed data seeds, and are shared by all runs. Values reported as mean $\pm$ sd use the population standard deviation over seeds; tests and confidence intervals use the usual sample estimates.

\textbf{Configuration (identical across arms).} $d_{\mathrm{model}} = 128$, 2 layers, 4 heads, 10{,}000 training samples, batch 64, AdamW ($lr = 5\times10^{-4}$, weight decay 0.01), 5 epochs, 5 paired seeds, CPU (PyTorch 2.8). Sequences are padded per batch; loss ignores padding.

\subsection{Main results}\label{sec:main}

\begin{table}[h]
\centering\footnotesize
\setlength{\tabcolsep}{4pt}
\resizebox{\textwidth}{!}{%
\begin{tabular}{lccccc}
\toprule
Attention mechanism & Curvature & OUT accuracy & OUT loss & Params & Training time$^{\dagger}$ \\
\midrule
Euclidean (dot product) & $0$ & 85.80\% $\pm$ 0.65 & 1.314 $\pm$ 0.028 & 413k & $1\times$ \\
Spherical FR exact (\S\ref{sec:spherical}) & $+1/4$ & 86.63\% $\pm$ 0.64 & 1.316 $\pm$ 0.031 & 413k & $\approx 1.1\times$ \\
Ricci--Mahalanobis (quadratic approx.)$^{*}$ & approx. & 86.87\% $\pm$ 0.71 & 1.298 $\pm$ 0.038 & 413k & $\approx 4\times$ \\
Hyperbolic anisotropic (\S\ref{sec:aniso}) & $-1/2$ per factor & 87.58\% $\pm$ 0.69 & 1.295 $\pm$ 0.054 & 479k & $28$--$110\times^{\ddagger}$ \\
\textbf{Hyperbolic isotropic (\S\ref{sec:iso}, ours)} & $<0$ & \textbf{88.43\% $\pm$ 0.77} & 1.309 $\pm$ 0.040 & \textbf{415k} & $\approx 1.4\times$ \\
Mixed-curvature adaptive (\S\ref{sec:mixed}) & learned & 86.90\% $\pm$ 0.72 & --- & 418k & $\approx 1.8\times$ \\
\bottomrule
\end{tabular}%
}
\caption{Dyck-2 out-of-distribution results (mean $\pm$ sd over 5 paired seeds). $^{*}$Quadratic midpoint-metric approximation from an earlier iteration of this project, retained as an ablation arm. $^{\dagger}$End-to-end training time relative to Euclidean attention in the configuration of Section~\ref{sec:protocol}. $^{\ddagger}$Attention-module micro-benchmark only, range over three machines (Section~\ref{sec:cost}, Appendix~\ref{app:cost}); on that micro-benchmark the isotropic kernel costs $2.3$--$2.8\times$.}
\label{tab:main}
\end{table}

\textbf{Within and beyond the training length.} The ranking is the same within the training lengths, with similar margins (IN accuracy: Euclidean $95.6\%$, spherical $96.2\%$, anisotropic $97.8\%$, isotropic $98.4\%$; isotropic minus Euclidean $+2.74$ pp IN and $+2.64$ pp OUT). Under this five-epoch budget the kernels therefore differ mainly in how well they learn the language; the advantage carries beyond the training length rather than arising there, and the Euclidean arm is not trained to convergence.

\begin{figure}[t]
\centering
\includegraphics[width=\textwidth]{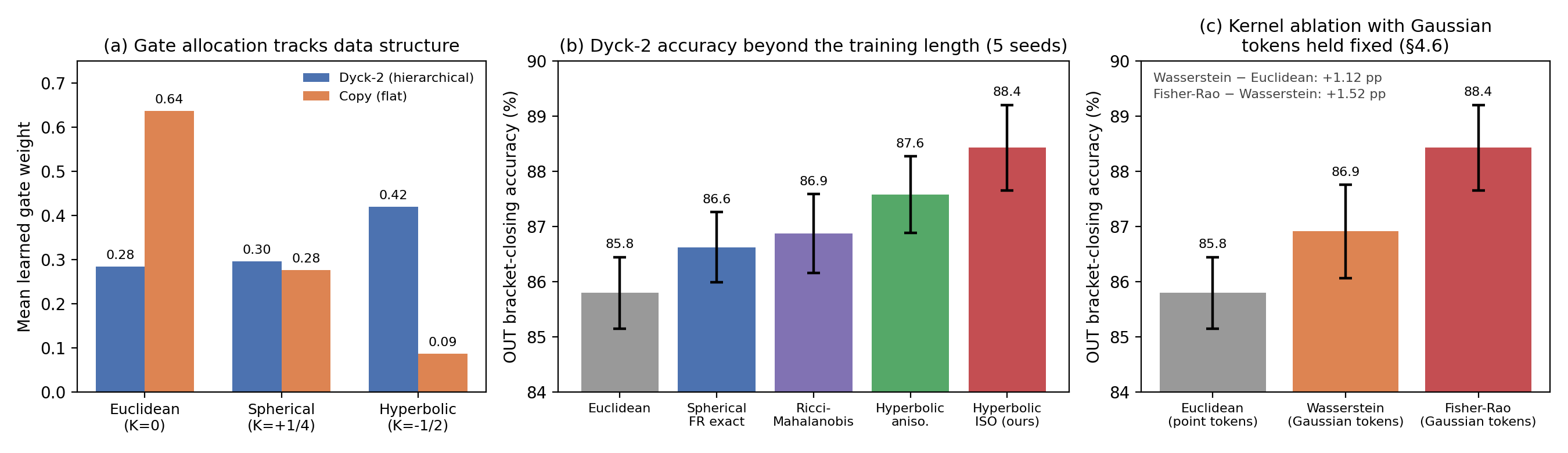}
\caption{(a) Learned gate allocation tracks data structure; (b) Dyck-2 accuracy beyond the training length by curvature (5 seeds, mean $\pm$ sd); (c) kernel ablation with Gaussian tokens held fixed: the Fisher--Rao kernel adds more over Wasserstein--Bures ($+1.52$ pp) than Gaussian tokens add over point tokens ($+1.12$ pp).}
\label{fig:results}
\end{figure}

\textbf{Accuracy ranks flat $<$ spherical $<$ hyperbolic} (Table~\ref{tab:main}, Figure~\ref{fig:results}b). The hyperbolic advantage is what the volume-growth argument for hierarchy embedding predicts \cite{nickel2017,gu2019}; the spherical kernel's small edge over flat attention ($p = 0.043$, uncorrected) is not predicted by it. The parameter-matched isotropic mechanism is the strongest arm: it wins all five seeds against every other arm except the anisotropic hyperbolic variant, which it beats in four of five (Section~\ref{sec:stats}).

\subsection{Statistical analysis}\label{sec:stats}

Paired two-sided $t$-tests over 5 shared seeds (df = 4) are given in Table~\ref{tab:contrasts}:

\begin{table}[h]
\centering\small
\begin{tabular}{lccc}
\toprule
Contrast (OUT accuracy) & $\Delta$ & $t$ & $p$ \\
\midrule
\textbf{Isotropic vs.\ Euclidean} & \textbf{+2.64 pp} & \textbf{+8.80} & \textbf{0.0009} \\
Isotropic vs.\ spherical & +1.81 pp & +4.60 & 0.010 \\
Isotropic vs.\ Ricci-approx. & +1.56 pp & +4.83 & 0.0085 \\
Isotropic vs.\ anisotropic & +0.85 pp & +2.55 & 0.063 \\
Anisotropic vs.\ Euclidean & +1.79 pp & +5.57 & 0.0051 \\
Anisotropic vs.\ spherical & +0.96 pp & +3.69 & 0.021 \\
Spherical vs.\ Euclidean & +0.83 pp & +2.92 & 0.043 \\
\bottomrule
\end{tabular}
\caption{Paired contrasts on Dyck-2 OUT accuracy.}
\label{tab:contrasts}
\end{table}

With Bonferroni correction across the seven contrasts of this table ($\alpha = 0.05/7 \approx 0.0071$), only the isotropic-vs-Euclidean and anisotropic-vs-Euclidean contrasts remain significant; the remaining contrasts are reported uncorrected. Beyond individual $p$-values, the ranking flat $<$ spherical $<$ hyperbolic (both hyperbolic arms) holds in every seed.

\textbf{The parameter confound is resolved by design:} the anisotropic arm carries 16\% extra parameters (479k), but the \emph{strongest} arm --- isotropic --- is parameter-matched to the Euclidean baseline (415k vs.\ 413k) and matches or exceeds the larger anisotropic model ($+0.85$ pp, $p = 0.063$, 4/5 seeds).

\subsection{Computational cost}\label{sec:cost}

Forward+backward of a single attention module (not a full training step), $B{=}32$, $L{=}60$, $d{=}128$, single CPU thread:

\begin{table}[h]
\centering\small
\begin{tabular}{lcc}
\toprule
Kernel & ms (fwd+bwd, one module) & relative \\
\midrule
Euclidean & 16.7 & $1\times$ \\
Hyperbolic isotropic (matmul decomposition) & 38.5 & $2.3\times$ \\
Log-Bhattacharyya approximation (5-D broadcast) & 348.4 & $21\times$ \\
Hyperbolic anisotropic exact (5-D broadcast) & 476.1 & $28.5\times$ \\
\bottomrule
\end{tabular}
\caption{Speed micro-benchmark (single attention module, forward+backward; original hardware).}
\label{tab:speed}
\end{table}

The decisive factor is not the transcendental function but tensor materialization: any per-coordinate pairwise formulation pays the $[B, H, L, L, d_h]$ broadcast; the isotropic reduction is what removes it. Because the projections and feed-forward layers are shared costs, the end-to-end training-time overhead is smaller than the module-level ratio: $\approx 1.4\times$ Euclidean in the configuration of Section~\ref{sec:protocol} (Sections~\ref{sec:main} and~\ref{sec:cat}). On two other machines (Appendix~\ref{app:cost}) the isotropic overhead stays at $2.5$--$2.8\times$, while the per-coordinate formulations range from $28\times$ to $110\times$, as expected when their cost is dominated by materialization.

\subsection{Ablations}\label{sec:ablations}

\textbf{Temperature.} The spherical kernel trains equally well with fixed unit temperature as with a learnable one (a learnable $\beta$ initialized at 8 drifted downward and slightly \emph{underperformed}, 86.2\% vs.\ 87.6\% single-seed) --- consistent with the intrinsic boundedness $d^2_{\mathrm{FR}} \in [0, \pi^2]$ supplying adequate logit range. The unbounded hyperbolic kernel does use a learnable temperature (one scalar per layer).

\textbf{Log-Bhattacharyya approximation.} Replacing the exact hyperbolic distance with the Gaussian log-Bhattacharyya coefficient (correct locally: the Bhattacharyya Hessian is $\tfrac14 g_F$) reached 87.9\% (single seed) but is both slower (Table~\ref{tab:speed}) and weaker than the exact isotropic kernel.

\textbf{Kernel shape: a positive-definite geodesic kernel.} The score $-\beta d^2$ makes each attention row a geodesic Gaussian kernel $\exp(-\beta d^2)$, which on a curved manifold is not positive definite for all $\beta$; the geodesic Laplacian kernel $\exp(-\beta d)$, in contrast, is positive definite on hyperbolic space \cite{feragen2015} (it is also the form of the matching function of hyperbolic attention networks on the hyperboloid \cite{gulcehre2019}, in its softmax variant). The distinction matters for scaling, because random-feature and linear-attention approximations require a positive-definite kernel. Replacing the score by $s_{ij} = -\beta\, \dFR(i,j)/\sqrt{d_h}$ --- same Gaussian tokens, same configuration (Section~\ref{sec:protocol}) and seeds --- gives $87.88\% \pm 1.16$ OUT accuracy: $+2.08$ pp over Euclidean attention (95\% CI $[0.91, 3.26]$, paired $t = 4.91$, $p = 0.008$, 5/5 seeds) and $0.55$ pp below the Gaussian form, a difference that is not significant ($p = 0.16$; 95\% CI for Fisher--Rao minus Laplacian $[-0.35, 1.46]$ pp). The positive-definite form thus retains most of the geometric advantage, which keeps a route open to linear-time variants of the kernel; equivalence with the Gaussian form is not established at $n = 5$, and no such approximation is tested here.

\subsection{Isolating the source of the gain: geometry vs.\ distributional tokens}\label{sec:wasserstein}

The closest prior mechanisms to our hyperbolic kernel represent tokens as Gaussians and score attention by the \emph{Wasserstein} (Bures) distance \cite{erick2023stochastic,fan2022stosa}, not the Fisher--Rao geodesic. This is a consequential distinction: for isotropic Gaussians the Wasserstein geometry is flat ($K = 0$; $W_2^2$ below is the squared distance of a flat metric on the same $(\mu, \sigma)$ coordinates), whereas the Fisher--Rao geometry is hyperbolic. We therefore ablate the \emph{geometry} while holding the \emph{token representation} fixed --- identical isotropic-Gaussian $\mu, \sigma$ projections and architecture, changing only the pairwise distance. For isotropic Gaussians the Bures distance also has matmul-friendly closed form $W_2^2 = \lVert\Delta\mu\rVert^2 + d_h\,\Delta\sigma^2$.

A second control separates the \emph{global} geometry of the Fisher--Rao kernel from its \emph{local} information geometry. The Kullback--Leibler divergence between the same isotropic Gaussians --- the Bregman divergence of the log-partition function --- agrees with the Fisher metric to second order, so it shares the kernel's local information geometry; but it grows quadratically with mean separation, whereas the hyperbolic geodesic grows logarithmically, which is how negative curvature shows up in these coordinates. It too reduces to one matrix multiplication: $\mathrm{KL}(q_i \Vert k_j) = \tfrac12\big[d_h\,\sigma_i^2/\sigma_j^2 + \lVert\Delta\mu_{ij}\rVert^2/\sigma_j^2 - d_h - d_h \ln(\sigma_i^2/\sigma_j^2)\big]$, used as $s_{ij} = -\beta\,\mathrm{KL}(q_i \Vert k_j)/d_h$. Table~\ref{tab:geomvsrep} gives the results.

\begin{table}[h]
\centering\footnotesize
\begin{tabular}{llc}
\toprule
Kernel (Dyck-2, 5 paired seeds) & Geometry & OUT accuracy \\
\midrule
Euclidean & flat, point tokens & 85.80\% $\pm$ 0.65 \\
Wasserstein--Bures (Gaussian tokens) & optimal transport, non-hyperbolic & 86.91\% $\pm$ 0.85 \\
KL / Bregman (Gaussian tokens) & local information geometry, non-hyperbolic & 87.32\% $\pm$ 0.77 \\
\textbf{Fisher--Rao (Gaussian tokens, ours)} & \textbf{hyperbolic} & \textbf{88.43\% $\pm$ 0.77} \\
\bottomrule
\end{tabular}
\caption{Geometry-vs-representation ablation, with the KL (Bregman) control for local information geometry.}
\label{tab:geomvsrep}
\end{table}

Using Gaussian tokens at all contributes $+1.12$ pp over the point-token baseline (Wasserstein vs.\ Euclidean, paired $t=3.07$, $p=0.037$); \emph{choosing the hyperbolic Fisher--Rao geometry rather than Wasserstein} contributes a further $+1.52$ pp (paired $t=3.33$, $p=0.029$) --- the larger share (Figure~\ref{fig:results}c). The KL control sharpens the attribution. With the same tokens and the same local information geometry, KL reaches $87.32\%$ --- above Euclidean attention ($+1.52$ pp, $p = 0.0006$) and not significantly different from Wasserstein ($+0.41$ pp, $p = 0.23$) --- but the Fisher--Rao kernel exceeds it by $+1.11$ pp (95\% CI $[0.44, 1.78]$, paired $t = 4.60$, $p = 0.010$, 5/5 seeds). What Fisher--Rao adds over KL is not the local metric, which the two share, but how the score grows with mean separation: logarithmically for the squared geodesic distance, quadratically for KL. That growth belongs to the link applied to the distance as much as to curvature: the squared Lorentzian distance used in model-space hyperbolic attention \cite{chen2022fully} is a function of the same hyperbolic distance yet grows quadratically in these coordinates. The KL control therefore isolates score growth, not curvature. The cleanest curvature contrast here is Fisher--Rao versus Wasserstein--Bures, which holds coordinates and link fixed and changes only the curvature ($+1.52$ pp, $p = 0.029$, $n = 5$); and a variant with per-token scales removed, whose score is a slowly growing function of the Euclidean distance between means, keeps most of the gain (Section~\ref{sec:sigma}). The gain is thus carried by the geodesic kernel --- its slow growth and, to an extent these controls do not separate, its curvature --- rather than by distributional tokens or by information geometry per se.

\subsection{Parameter and compute efficiency: matching accuracy at lower dimension}\label{sec:efficiency}

Sections \ref{sec:main}--\ref{sec:wasserstein} compare geometries at matched $d_{\mathrm{model}}$, which matches parameter count but leaves open a different question, closer in spirit to the original motivation for hyperbolic embeddings \cite{nickel2017}: since hyperbolic space packs exponentially more distinguishable structure into a fixed number of dimensions than Euclidean space, can the hyperbolic-isotropic kernel reach a \emph{given} accuracy using \emph{fewer} dimensions --- i.e., fewer parameters and less compute --- than Euclidean attention needs to reach the same accuracy? We sweep $d_{\mathrm{model}} \in \{16, 32, 64, 128, 256\}$ ($n_{\mathrm{heads}}=4$ fixed, so $d_h \in \{4,8,16,32,64\}$), 5 paired seeds --- with 2{,}000 training sequences and 400-sequence evaluation sets, otherwise the configuration of Section~\ref{sec:protocol}, so absolute accuracies sit below those of Section~\ref{sec:main} --- and compare each geometry against the \emph{other} geometry at \textbf{double the dimension} (Table~\ref{tab:efficiency}).

\begin{table}[h]
\centering\footnotesize
\setlength{\tabcolsep}{4pt}
\begin{tabular}{lccccc}
\toprule
Comparison & FR params & Eucl.\ params & Ratio & $\Delta$ accuracy & Wall-clock (5 ep.)$^{*}$ \\
\midrule
FR@$d_h{=}4$ vs.\ Eucl.@$d_h{=}8$ & 8.9k & 29.6k & 3.3$\times$ & $-2.59$ pp ($p=0.0047$) & $\approx$4 s vs.\ $\approx$2 s \\
FR@$d_h{=}8$ vs.\ Eucl.@$d_h{=}16$ & 30.1k & 108.3k & 3.6$\times$ & $-1.53$ pp ($p=0.0020$) & $\approx$4 s vs.\ $\approx$3 s \\
\textbf{FR@$d_h{=}16$ vs.\ Eucl.@$d_h{=}32$} & \textbf{109.3k} & \textbf{413.2k} & \textbf{3.8$\times$} & $+0.22$ pp ($p=0.27$) & \textbf{$\approx$4 s vs.\ $\approx$4 s} \\
\textbf{FR@$d_h{=}32$ vs.\ Eucl.@$d_h{=}64$} & \textbf{415.3k} & \textbf{1612.8k} & \textbf{3.9$\times$} & $-0.11$ pp ($p=0.74$) & \textbf{$\approx$6 s vs.\ $\approx$9 s} \\
\bottomrule
\end{tabular}
\caption{Cross-dimension efficiency sweep: Fisher--Rao at dimension $D$ vs.\ Euclidean at $2D$. $^{*}$Per 5-epoch run, recorded in whole seconds.}
\label{tab:efficiency}
\end{table}

At small head width ($d_h \le 8$) the hyperbolic kernel does \textbf{not} achieve this: it is significantly \emph{worse} than Euclidean attention at double the dimension, so no efficiency claim holds in this regime. Above a threshold ($d_h \ge 16$), the pattern inverts: Fisher--Rao at dimension $D$ shows no significant accuracy difference from Euclidean at dimension $2D$ (paired $t$-test, $p = 0.27$ and $p = 0.74$; 95\% CIs of the difference $[-0.26, +0.71]$ and $[-0.98, +0.76]$ pp --- no detectable difference at $n = 5$, although a deficit of up to $\approx 1$ pp cannot be excluded) while using $\approx 3.8$--$3.9\times$ fewer parameters. At the largest scale tested this is not just a parameter count artifact: real wall-clock training time is also lower for Fisher--Rao ($\approx 6$ s vs.\ $\approx 9$ s, roughly $1.5\times$ faster) because the $O(d_{\mathrm{model}}^2)$ projection cost that dominates at these widths scales quadratically with the doubled Euclidean dimension, outweighing the constant-factor $\arccosh$ overhead of the geodesic kernel (Section~\ref{sec:cost}). We read this as a genuine, if narrow, efficiency result: \emph{at sufficient scale, encoding tokens as points on a curved statistical manifold lets comparable accuracy be reached with a smaller latent width} --- echoing, in the attention setting, the dimension-efficiency argument originally made for static hyperbolic embeddings \cite{nickel2017}. Three qualifications keep this claim honest. First, the result is \emph{comparable accuracy at lower resource cost}, not higher accuracy: the practical implication is a smaller model, not a better one. Second, the comparison is between two different mechanisms (the hyperbolic kernel carries small additional $\sigma$ projections and the $\arccosh$), not a controlled scaling of one mechanism, so ``fewer parameters'' bundles the geometry with minor architectural differences; the parameter gap ($3.8$--$3.9\times$) is however far larger than those differences ($<1\%$ of either model). Third, the threshold behavior is a real limitation: this is not ``hyperbolic is always more efficient,'' it is ``hyperbolic becomes more efficient once each head has enough room to represent a mean vector and a meaningful scale,'' the mechanism behind the crossover at $d_h \approx 16$ is not explained by the present experiments --- a pre-registered probe of one candidate, that narrow heads operate in an effectively flat regime, did not account for it (even at $d_h = 4$ the trained kernel operates beyond the curvature radius; protocol and data in the repository) --- and no claim is made about where the threshold falls outside this architecture and task.

\subsection{Extended evidence: statistical power, model scale, depth stratification, and natural data}\label{sec:extended}

In response to the evidentiary concerns of Section~\ref{sec:limitations} (few seeds, small models, synthetic-only data, absolute positional embeddings), we ran seven follow-up experiments. Some strengthen the central claim substantially, others bound it; all seven are reported with equal weight.

\textbf{(a) Twenty seeds.} The headline contrast (hyperbolic-isotropic vs.\ Euclidean, Section~\ref{sec:main} configuration) extended from 5 to 20 paired seeds: $\Delta = +2.10$ pp (95\% CI $[1.74, 2.46]$), paired $t = +12.21$, $p = 1.9\times10^{-10}$, with the hyperbolic kernel winning \textbf{20 of 20 seeds}. The main result is not a small-sample artifact.

\textbf{(b) The advantage is larger in one $46\times$ configuration.} At $46\times$ the parameter count (6 layers, $d_{\mathrm{model}} = 512$, 8 heads, 19.0M parameters; 3 paired seeds), the gap \emph{widens} rather than evaporating: OUT accuracy $87.4\% \pm 0.3$ (Euclidean) vs.\ $92.1\% \pm 0.6$ (hyperbolic), $\Delta = +4.72$ pp (paired $t = 8.72$, $p = 0.013$, 3/3 seeds) --- more than double the $+2.1$ pp observed at the small configuration. It argues for evaluation at genuine scale (GPU, natural-language corpora) but does not establish a scaling trend: three seeds, one configuration trained on the same 10k sequences with the same learning rate, and in-distribution accuracy was not saved for this run.

\textbf{(c) The advantage does \emph{not} grow with hierarchy depth (pre-registered negative).} On Dyck-4 (4 bracket types, 4-layer models, training lengths 50--100, extrapolation 200--300), we stratified closing accuracy by the stack depth at each prediction, with the success criterion written before execution ($\Delta_{\mathrm{deep}} > +5$ pp at $p<0.01$; H1: $\Delta$ increasing in depth). The hyperbolic kernel beat Euclidean at \textbf{every} depth (shallow $\le 4$: $+3.05$ pp, $p=0.0013$; mid 5--8: $+3.09$ pp, $p=0.0044$; deep $\ge 9$: $+1.64$ pp, $p=0.017$; $n=5$, bins of 46k--75k predictions), but the advantage \emph{shrinks} rather than grows at depth --- \textbf{H1 is refuted and the pre-registered threshold was not met.} The mechanism is a broad formal-sequence improvement, not a deep-hierarchy specialist; the intuitive ``more curvature helps more nesting'' scaling story is not supported. A hybrid architecture (hyperbolic early layers, Euclidean late, motivated by the Section~\ref{sec:gate} gate observations) was also \emph{worse} than pure hyperbolic at depth ($\Delta = -1.62$ pp, $p = 0.010$), refuting the layer-specialization design hypothesis as well.

\textbf{(d) Program source: a diagnosis-driven norm-capped variant degrades less beyond the training length, after two failed attempts with the original kernel.} On character-level modeling of real Python source (the standard library of CPython 3.9.6, split by file: 1.20M characters train, 37.8k held out in 4 files; 97-symbol vocabulary including padding; each model sees 4,000 windows of 128 characters for 3 epochs, batch 32, and learned absolute positions extend to 512, so rows 128--511 are never trained), the kernel is consistently \emph{worse} in-distribution (train and test at equal length: a dedicated 3-seed run at $L{=}256$, $\Delta = +0.040$ bpc, $p = 0.09$, 3/3 seeds; and the in-distribution $L{=}128$ control of each of the three extrapolation runs below, all significant in Euclidean's favor). The test that the formal-language results motivate is the one beyond the training length: train on 128-character sequences, evaluate on the portion of longer sequences beyond the trained length. We ran this test twice with pre-registered endpoints (primary: extrapolative-zone bpc at $4\times$ training length, $L{=}512$, $p<0.05$). \textbf{With the original kernel, the pre-registered endpoint failed both times} (original $n{=}5$: $\Delta = -0.077$ bpc in the kernel's favor, $p = 0.136$; independent replication $n{=}10$, seeds new to the code task: $\Delta = -0.049$, $p = 0.28$), while the \emph{secondary} endpoint --- the extrapolative zone at $2\times$ training length --- favored the kernel significantly in both samples ($p = 0.027$, $p = 0.037$), and the failure mode at $4\times$ was identified as \emph{variance}, not mean: seed-to-seed spread $2.0$--$2.4\times$ Euclidean's, with some seeds excellent and some degrading.

We then diagnosed the instability directly. Instrumenting good and bad seeds (a first hypothesis --- $\sigma$ drift at unseen positions --- was tested and \emph{refuted}: $\sigma$ statistics do not separate good from bad seeds), the discriminating variable proved to be \textbf{outlier $\mu$ norms at untrained positions} (max $\lVert\mu\rVert$ 15--17 in failing seeds vs.\ 11--13 in good seeds, amplified by slightly larger learned $\beta$): under out-of-distribution positional inputs, unbounded $\mu$ projections occasionally throw tokens far out on the manifold, distorting the distance distribution. The minimal targeted fix is a \textbf{$\mu$-norm ball cap} ($\lVert\mu\rVert \le 12$, chosen from the diagnostic seeds before any confirmatory run) --- the direct analogue of the ball-clamping universally used in Poincar\'e implementations.

\textbf{The stabilized kernel passed the pre-registered endpoint on the third and final attempt} ($n{=}20$ seeds new to the code task, 16--35): extrapolative-zone bpc at $L{=}512$ of $3.738 \pm 0.094$ vs.\ Euclidean $3.822 \pm 0.097$, $\Delta = -0.084$, paired $t = -3.30$, $\bm{p = 0.0037}$, 15/20 seeds, with the pre-specified stability criterion also met (seed-to-seed spread at most $1.5\times$ Euclidean's; observed $0.97\times$). The moderate-extrapolation zone was significant for the third consecutive independent sample ($\Delta = -0.142$, $p < 10^{-4}$); the in-distribution control remains in Euclidean's favor, as in all prior runs. We state the epistemic status precisely: the \emph{original} kernel does not meet the far-extrapolation endpoint (two pre-registered failures); the \emph{stabilized} variant --- one additional clamp, motivated by a documented failure-mode diagnosis on separate seeds --- meets it in one adequately powered pre-registered confirmation. Two fair objections deserve direct answers. \emph{Is the cap theory-derived?} No --- it is an engineering safeguard, exactly analogous to the ball-clamping every practical Poincar\'e implementation uses, and we present it as such. \emph{Is $c = 12$ a tuned magic number?} A sensitivity analysis on five further fresh seeds shows the effect direction and magnitude persist across a $2\times$ range of the cap ($c = 8$: $\Delta = -0.111$; $c = 16$: $\Delta = -0.119$; 4/5 seeds each, descriptive $n=5$) --- the stabilization is a robust family, not a point estimate. All three confirmatory attempts and the sensitivity analysis are reported; replication at larger scale is the natural next test.

\textbf{(e) Technical English prose: the same crossed pattern at the first attempt.} Code is natural data but not natural language, so we repeated the design of (d) on a corpus not used in any earlier experiment of this project: technical English prose rendered from the local system manual (274 man pages of section 1; 2.73M characters train / 0.42M test, split by page; 119-symbol vocabulary including padding). Everything else was held identical to (d) --- character-level modeling, training at 128 characters, extrapolative-zone evaluation at $2\times$ and $4\times$ training length, the same architecture and budget --- with the stabilized kernel against Euclidean attention, ten fresh seeds (46--55), and the endpoint written before execution (extrapolative-zone bpc at $L{=}512$, $p<0.05$). \textbf{The pre-registered endpoint was met at the first attempt:} $3.965 \pm 0.036$ vs.\ Euclidean $4.023 \pm 0.032$ bpc, $\Delta = -0.058$, paired $t = -6.02$, $\bm{p = 0.0002}$, \textbf{10/10 seeds}; the stability criterion was also met (seed spread $1.14\times$ Euclidean's, below the pre-specified $1.5\times$). The moderate-extrapolation zone ($L{=}256$) agrees ($\Delta = -0.088$, $p = 0.0002$), and the in-distribution control again favors Euclidean attention, exactly as declared in advance ($\Delta = +0.032$ bpc, $p < 10^{-4}$). This is the second natural-data family --- after program source --- in which the same crossed pattern (worse in distribution, better beyond the training length) appears under pre-registered endpoints; on formal languages, by contrast, the kernel leads both within and beyond the training length (Sections~\ref{sec:main} and~\ref{sec:stress}). Its scope is deliberately narrow: technical prose (manual pages are terse and repetitive relative to narrative or conversational text), character level, $\approx$495k parameters (the 119-symbol vocabulary and 512-position table make it larger than the model of Section~\ref{sec:main}), CPU; general-domain text at token level and at scale remains untested. For scale, a smoothed character bigram reaches $3.74$ bpc and a 4-gram $2.42$ bpc on the same test text (additive smoothing with constant 0.05; \texttt{verificacion\_claude/}\allowbreak\texttt{ngram\_referencia.py}), against $3.60$--$3.64$ for both arms in distribution: the models are near character-bigram quality, and the comparison measures how two under-trained models degrade beyond the training length, not natural-language extrapolation.

\textbf{(f) Relative positions: the advantage does not depend on absolute positional embeddings.} Every result above uses learned absolute positional embeddings, so positions beyond the training length enter as vectors that were never trained, and part of what ``length extrapolation'' measures is robustness to them. We therefore removed absolute positions from both arms and gave each a relative scheme: rotary embeddings (RoPE) \cite{su2021roformer} on queries and keys for Euclidean attention, and the same rotation applied to $\mu_q$ and $\mu_k$ for the Fisher--Rao kernel. For the latter each rotation fixes $\sigma$ and acts on $\mathbb{H}^{d+1}$ as an isometry (a rotation about the vertical geodesic $\mu = 0$), so rotated tokens stay on the manifold and the score remains an exact Fisher--Rao distance; since $R_i^{\top}R_j = R_{j-i}$, $d(R_i\mu_q, R_j\mu_k) = d(\mu_q, R_{j-i}\mu_k)$ depends on positions only through $j - i$, and the single-matmul decomposition is preserved. The test was pre-registered, with the protocol hashed before execution: 20 paired seeds fixed in the protocol (6--25), the configuration of Section~\ref{sec:protocol}, Dyck-2 OUT accuracy as primary endpoint, and the copy task of Section~\ref{sec:gate} as secondary (Table~\ref{tab:rope}).

\begin{table}[h]
\centering\footnotesize
\setlength{\tabcolsep}{4pt}
\begin{tabular}{lccccc}
\toprule
Task (RoPE, both arms) & Euclidean & Fisher--Rao & $\Delta$ (95\% CI) & $p$ & Seeds won \\
\midrule
\textbf{Dyck-2, OUT (primary)} & 90.91\% $\pm$ 1.10 & \textbf{92.73\% $\pm$ 1.09} & \textbf{+1.82 pp} [1.08, 2.57] & $6\times10^{-5}$ & 16/20 \\
Dyck-2, IN & 95.76\% & 97.42\% & +1.66 pp & $7\times10^{-8}$ & --- \\
\textbf{Copy, OUT (secondary)} & 37.85\% $\pm$ 2.97 & \textbf{43.44\% $\pm$ 5.00} & \textbf{+5.60 pp} [3.09, 8.10] & $0.0002$ & 16/20 \\
Copy, IN & 90.09\% & 92.88\% & +2.79 pp & $6\times10^{-13}$ & --- \\
\bottomrule
\end{tabular}
\caption{Pre-registered relative-position test: rotary position embeddings (RoPE) in both arms, mean $\pm$ sd over 20 paired seeds (6--25); 95\% CI of the paired difference in brackets.}
\label{tab:rope}
\end{table}

\textbf{The pre-registered endpoint was met.} Relative positions help both arms substantially --- Euclidean attention with RoPE (90.91\%) exceeds even the Fisher--Rao kernel with absolute positions (88.43\%, Section~\ref{sec:main}), which confirms that absolute embeddings were a real handicap --- and the Fisher--Rao kernel keeps its lead under the fairer scheme ($+1.82$ pp, paired $t = 5.13$, $p = 6\times10^{-5}$). The shape of the lead is informative. On Dyck-2 it is the same within and beyond the training length ($+1.66$ vs.\ $+1.82$ pp; the drop from IN to OUT is 4.85 pp for Euclidean attention and 4.69 pp for Fisher--Rao; post hoc paired $t$, $p = 0.60$): the kernel learns the hierarchical task better overall, and that advantage carries intact to unseen lengths rather than adding a separate extrapolation benefit. On the copy task the lead does grow beyond the training length ($+2.79$ pp within, $+5.60$ pp beyond; the Fisher--Rao kernel degrades 2.8 pp less; post hoc paired $t$, $p = 0.025$), and copy extrapolation improves over the near-chance $\approx$30\% of absolute positions (Section~\ref{sec:gate}), although it remains far from solved for both arms (38--43\%). With fair positions, then, the case for the kernel rests on better learning of structured sequences, an advantage that persists --- and on copy partly grows --- beyond the training length.

\textbf{(g) A boundary: the pattern does not appear on Spanish clinical text.} To test whether the crossed pattern of (d)--(e) extends to another language and domain, we repeated the design of (e) on Spanish clinical prose: the evidence summaries of the evidence-to-decision frameworks of 82 publicly available Chilean national clinical practice guidelines (5.55M training and 0.91M test characters, split by guideline so that no test guideline contributes training text; 250-symbol vocabulary including padding), with 20 fresh seeds (201--220) and the endpoint written and hashed before training. \textbf{The pre-registered endpoint was not met:} extrapolative-zone bpc at $L{=}512$ of $4.051 \pm 0.022$ vs.\ Euclidean $4.047 \pm 0.022$, $\Delta = +0.004$ (95\% CI $[-0.007, +0.014]$, which excludes effects of the size seen on code and English prose), paired $t = 0.78$, $p = 0.45$, 8/20 seeds. The moderate-extrapolation zone shows no advantage either ($\Delta = +0.008$, $p = 0.11$), the in-distribution control again favors Euclidean attention, as predicted ($\Delta = +0.017$ bpc, $p < 10^{-4}$), and the stability criterion is met (spread $0.97\times$ Euclidean's). On this corpus the kernel keeps its in-distribution cost without the extrapolation benefit. The test does not tell which difference matters --- language, domain, the larger vocabulary, or the templated passages that recur across guidelines --- but it bounds the natural-data claim to the two families in which the pattern was observed.

\textbf{Synthesis.} The picture these results paint is more specific than ``hyperbolic helps hierarchies'': the advantage is (i) robust across seeds, (ii) larger in one $46\times$ configuration, (iii) present across all hierarchy depths but front-loaded at shallow--mid depth rather than deep, and (iv) not reproduced in distribution on natural data, where small models near character-bigram quality show a different, crossed pattern: with untrained absolute positions the norm-capped variant degrades less beyond the training length on program source --- once a diagnosed instability (unbounded $\mu$ projections at unseen positions) is corrected by a one-line ball cap --- and on English prose, though not on Spanish clinical text (Section~\ref{sec:extended}g); and (v) not tied to absolute positional embeddings: on Dyck-2 and copy it persists with relative (rotary) positions, on Dyck-2 as a general advantage on hierarchical structure rather than an extrapolation-specific one (Section~\ref{sec:extended}f). In-distribution natural-data modeling remains consistently in Euclidean attention's favor.

\subsection{What the learned scale encodes: nesting depth, with opposite query and key codes}\label{sec:sigma}

The isotropic kernel gives every token, in every head, two learned scales: $\sigma_q$ as a query and $\sigma_k$ as a key (Section~\ref{sec:iso}). Point tokens have no such parameters: the scales enter with the isotropic Gaussian family, and so does the question of what they learn. The geometry invites reading them as heights in the half-space (Figure~\ref{fig:halfspace}), but nothing in training tells the model what they should mean. We asked what they actually learn in six pre-registered protocols with nine primary endpoints, mostly on Dyck-2 in the configuration of Section~\ref{sec:protocol} (Table~\ref{tab:sigma}). Each protocol was hashed (SHA-256) before its full run, after the smoke test recorded in its seal, and, except the intervention test, which acts on the models of the first test, used initialization seeds not used before; all tests share the same training and evaluation sequences.

\begin{table}[h]
\centering\footnotesize
\setlength{\tabcolsep}{3pt}
\resizebox{\textwidth}{!}{%
\begin{tabular}{llll}
\toprule
Test (seeds) & Pre-registered endpoint & Result & Met \\
\midrule
Depth in $\sigma$ (401--405) & $S \ge 0.10$ and above untrained, $p<0.05$ & $S = 0.584$ vs.\ $0.001$, $p = 1.5\times10^{-4}$ & yes \\
Opposite directions, layer 2 (601--605) & $\sigma_q$ slope $<0$ and $\sigma_k$ slope $>0$ in $\ge 18/20$ & 20/20 and 20/20 (all heads, 5/5 seeds) & yes \\
Other hierarchies (801--805) & same signs in $\ge 16/20$ & Dyck-4: 20/20, 20/20; deep Dyck-2: 16/20, 20/20 & yes \\
Depth removed from $\sigma_q$ (401--405) & larger drop without depth, $p<0.05$ & $+0.15$ pp, $p = 0.038$ & yes$^{\dagger}$ \\
Fixed $\sigma$ in layer 1 (701--710) & non-inferior, margin $-0.5$ pp & $+0.04$ pp $[-0.19, +0.28]$ & yes \\
$\sigma_q \leftrightarrow \sigma_k$ swap (801--805) & OUT drop $>0$, $p<0.05$ & $4.3$ pp, 5/5 seeds, $p = 0.051$ & no \\
$\sigma_k$ as error signal (601--605) & AUC $\ge 0.70$, CI above 0.5 & $0.835$ ($0.53$ adjusted) & yes$^{\ddagger}$ \\
Depth in $\sigma$, Python source (501--505) & $S \ge 0.05$ and above untrained, $p<0.05$ & $S = 0.026$, $p = 0.056$ & no \\
Keys track their pair's $\sigma'^{*}$ (801--805) & partial $\rho>0$ in $\ge 16/20$, above a non-pair control & 12/20 (13/20 above control) & no \\
\bottomrule
\end{tabular}%
}
\caption{Pre-registered tests of what the per-token scale learns (isotropic kernel; Dyck-2 unless stated; $n=5$ seeds, $n=10$ for non-inferiority; 20 cases $=$ 4 second-layer heads $\times$ 5 seeds). $S$: largest gain in $R^2$ from adding stack depth to a regression of $\log\sigma$ on symbol and position, over the 16 layer--head--role combinations. The 20 cases are not independent (4 heads per seed). $^{\dagger}$Below the 1 pp drop that the protocol fixed in advance as the level under which there is no evidence that the model depends on $\sigma_q$. $^{\ddagger}$Endpoint met, but the signal is not independent of position, depth and symbol (text). Each endpoint was tested at $\alpha = 0.05$ without correction; under a Holm correction across the nine, the intervention result would not survive.}
\label{tab:sigma}
\end{table}

\textbf{Depth is encoded, in the second layer.} For every token we regressed $\log\sigma$ on its symbol and ten position bins and measured the gain in $R^2$ from adding its stack depth, separately for each layer, head and role; the per-seed statistic $S$ is the largest of the 16 gains. Trained models reach $S = 0.584$, against $0.001$ for the same models before training (paired $t(4) = 14.09$, $p = 1.5\times10^{-4}$), and the combination selected in distribution keeps $S = 0.484$ on longer sequences (lengths 80--120) without reselection. That combination is the second-layer query scale in 5/5 seeds. First-layer scales carry no depth information ($\Delta R^2 \approx 0$), as expected: the first layer sees only symbols and positions. The untrained model is a weak null: in a trained network whose residual stream must carry depth, other one-dimensional readouts may also correlate with it, so $S$ alone does not show that $\sigma$ is specifically tuned to depth; the opposite-sign pattern below is the more specific evidence.

\textbf{Queries narrow and keys widen.} In the second layer the two roles move in opposite directions with depth. On fresh seeds the depth slope of $\log\sigma_q$ is negative and that of $\log\sigma_k$ positive in all four second-layer heads of every seed (20/20 head--seed pairs each). The heads within a seed are not independent, so the seed-level evidence is 5/5 (one-sided sign test $p = 0.031$), with mean slopes of $-0.372$ (95\% CI $[-0.423, -0.321]$) and $+0.167$ ($[0.154, 0.180]$) per level. Each level of nesting multiplies $\sigma_q$ by $\approx 0.69$ and $\sigma_k$ by $\approx 1.18$; in isometry-invariant terms, the vertical separation $\ln(\sigma_k/\sigma_q)$ grows by $\approx 0.54$ per level. These factors are log-linear summaries: $\sigma_q$ falls mostly over the first levels and then saturates, while $\sigma_k$ grows almost linearly. The pattern holds on longer sequences (40/40 head--seed--role cases, at 0.5--0.6 of the in-distribution slope) and is absent before training (slopes two orders of magnitude smaller). It is not specific to two bracket types: on further fresh seeds it reappears on a four-type variant (Dyck-4) generated with the same depth sequences (20/20 and 20/20; $\times 0.70$ and $\times 1.21$ per level) and on a deeper Dyck-2 distribution whose sequences reach a median maximum depth of 10 (16/20, at the pre-registered threshold, and 20/20; $\times 0.89$ and $\times 1.13$; the weaker query slope fits the saturation). Queries thus follow the geometric reading of $\sigma$ --- deeper, closer to the boundary --- while keys invert it.

\textbf{Removing the code at test time costs little.} Replacing the second-layer $\sigma_q$ by a linear surrogate fitted on the token's symbol alone (which removes depth and position; $R^2$ with the original $\log\sigma_q$ of 0.28) lowers out-of-distribution closing accuracy by 0.15 pp more than a surrogate fitted on symbol, position and depth ($R^2$ 0.71; $t(4) = 3.04$, $p = 0.038$, 4/5 seeds; 0.41 pp in distribution, $p = 0.001$); the depth-specific contrast (symbol and position vs.\ symbol, position and depth) is 0.09 pp ($p = 0.040$; 0.28 pp in distribution). The pre-registered endpoint is met, but the drop is below the 1 pp that the protocol fixed in advance as the level under which there is no evidence that the model depends on $\sigma_q$, and it would not survive a correction across the nine endpoints. The depth code is thus strongly decodable but barely used at test time; depth presumably reaches the prediction through channels we did not intervene on, in line with the dissociation between decodable and causally used depth reported for bracket-sequence transformers \cite{sharma2026}. Exchanging $\sigma_q$ and $\sigma_k$ at test time lowers out-of-distribution accuracy in all five seeds (mean 4.3 pp, range 1.7--10.3 pp), but the pre-registered test was not significant ($t(4) = 2.75$, $p = 0.051$), so we make no claim; secondary, uncorrected comparisons point the same way (swap vs.\ equalizing the two scales, 2.6 pp, $p = 0.004$; Dyck-4, 5.9 pp, $p = 0.0008$). The swap also exchanges each role's overall level of $\sigma$, so it does not isolate the direction of the depth code. Per-token scales are needed only where the code lives: fixing $\sigma$ in the first layer to one learned scalar per head and role is non-inferior to per-token scales (difference $+0.04$ pp, 95\% CI $[-0.19, +0.28]$, margin $-0.5$ pp, about a fifth of the gap over dot-product attention; $n = 10$), whereas fixing it in both layers costs 0.71 pp ($p = 0.004$). Two consequences follow. In the first layer the two fixed scales end up equal (starting equal, they receive equal gradients), so a fixed first-layer scale is a single per-head constant that $W_\mu$ can absorb. And with $\sigma$ fixed in both layers the score is a slowly growing function of the Euclidean distance between means, $2\sqrt{2}\,\mathrm{arcsinh}\big(\lVert\Delta\mu\rVert/(2\sqrt{2}\,\sigma)\big)$ (tokens on a horosphere), yet this variant still reaches $87.69\%$, above the $85.80\%$ of dot-product attention in Table~\ref{tab:main} (an unpaired comparison on different seeds): most of the Dyck-2 gain does not need per-token scales. This matches constructions in which a first layer only counts depth, with uniform attention, and a second layer matches by depth \cite{yao2021}.

\textbf{What the code is not, and what remains open.} The second-layer key scale flags closing errors (AUC 0.835 in distribution, above the pre-registered 0.70; 0.687 out of distribution, below it), but mostly by tracking position, depth and symbol: position alone reaches 0.823, adjusted for the three its AUC falls to 0.53 (interval above 0.5 in 3/5 seeds), and adding it to a cross-validated logistic model of those variables changes the AUC by $+0.00006$ on average. Inside the kernel, as outside it (Section~\ref{sec:limitations}), $\sigma$ is not an independent error or uncertainty signal. On real Python source (the stabilized kernel of Section~\ref{sec:extended}d, with indentation as depth), the pre-registered depth code was not detected at the threshold ($S = 0.026$ against 0.05; $p = 0.056$ against the untrained model, which the trained model exceeds in 5/5 seeds; expected direction in 2/5 seeds), so a small signal is not excluded. Finally, why keys widen is not explained. For a query of scale $\sigma$ and a key mean at separation $\lVert\Delta\mu\rVert$, the key scale that minimizes the distance is $\sigma'^{*} = \sqrt{\sigma^2 + \lVert\Delta\mu\rVert^2/2} \ge \sigma$ (the foot of the perpendicular from the query to the vertical geodesic above the key's mean), which suggested that keys widen to remain reachable by the queries that should find them. The data do not support it: trained keys are about $18\times$ narrower than $\sigma'^{*}$ (median ratio 0.054), and the scale of an opening bracket's key tracks the $\sigma'^{*}$ of its own closing query in only 12/20 head--seed pairs, although attention gives the matching opener more weight than the average other opener (20/20 heads; median 0.16 vs.\ 0.06). The data place the model in a far regime, where $\dFR \approx \sqrt{2}\,[\ln(\tfrac{1}{2}\lVert\Delta\mu\rVert^2 + \Delta\sigma^2) - \ln\sigma_q - \ln\sigma_k]$: to leading order $\ln\sigma_k$ shifts a key's distance to every query by the same amount, which a pair-specific test could not detect, and through the square in the score $\ln\sigma_q$ acts as a query-specific sharpness. Reading the code as ``query scale $\approx$ sharpness, key scale $\approx$ key-specific bias'' is a hypothesis we have not tested.

\textbf{Relation to prior work.} Encoding generality or hierarchical level in the scale of a representation is well established for static embeddings --- Gaussian variance \cite{vilnis2015}, whose Fisher geometry Tifrea et al.\ identify with a product of hyperbolic half-planes \cite{tifrea2019}, and hyperbolic norm \cite{nickel2017} --- and syntactic or stack depth is decodable from transformer representations \cite{hewitt2019,sharma2026}. Separate query and key variances also exist in distribution-based attention \cite{erick2023stochastic,fan2022stosa}, but we found no analysis of what they learn. What these tests add is evidence from inside the attention kernel, including an opposite-direction query/key scale code that we did not find reported in prior work (a targeted general web search on 29 October 2026, English only, without Google Scholar or Semantic Scholar). Their scope is narrow: synthetic bracket languages, a two-layer model, five seeds per test (ten for non-inferiority), CPU.

\section{Data-Driven Geometry Selection}\label{sec:gate}

The mixed-curvature mechanism (Section~\ref{sec:mixed}) was trained, with identical hyperparameters and \textbf{no geometric supervision of any kind}, on two tasks of opposite structure: Dyck-2 (hierarchical) and a flat copy task (first half random tokens, second half an exact repetition --- pure positional lookup, no hierarchy). Table~\ref{tab:gates} and Figure~\ref{fig:results}a report the learned allocations.

\begin{table}[h]
\centering\small
\begin{tabular}{lccc}
\toprule
Task & Euclidean gate & Spherical gate & Hyperbolic gate \\
\midrule
Dyck-2 (hierarchical) & 0.28 & 0.30 & \textbf{0.42} \\
Copy (flat) & \textbf{0.64} & 0.28 & 0.09 \\
\bottomrule
\end{tabular}
\caption{Learned gate allocation (mean over positions, heads, layers, seeds).}
\label{tab:gates}
\end{table}

The allocation tracks the structure of the data: hyperbolic-dominant under hierarchy, hyperbolic nearly silenced (0.09) under flat structure with the weight migrating to the Euclidean kernel. Per-layer resolution sharpens the picture: on the copy task the second layer is predominantly Euclidean (mean Euclidean gate 0.77 over 3 seeds, 0.53--0.98 per seed; in one seed all four heads reach 0.96--0.99), while on Dyck-2 the first layer specializes toward hyperbolic (mean hyperbolic gate 0.58 over 5 seeds, 0.51--0.64 per seed; single heads up to 0.85) with a more Euclidean second layer (mean Euclidean gate 0.41 vs.\ 0.16 in the first) --- suggesting a division of labor in which early layers extract hierarchical structure in negative curvature and later layers aggregate in flat geometry, although an architecture that hard-codes this division underperformed pure hyperbolic attention (Section~\ref{sec:extended}c).

\textbf{Caveats, stated plainly.} (i) On the purely hierarchical task the adaptive mixture (86.9\%) does not beat the pure hyperbolic specialist (88.4\%): when the data is uniformly hierarchical, mixing dilutes. (ii) Copy-task OUT extrapolation collapses to $\approx$30\% for \textbf{all} mechanisms because absolute positional embeddings do not extrapolate --- a positional confound, not a geometric one; copy-task claims are therefore restricted to gate behavior and in-distribution accuracy (where, incidentally, the isotropic kernel also leads: 83--91\% vs.\ 62\% Euclidean). With relative positions copy extrapolation improves ($\approx$30\% $\to$ 38--43\%) but remains far from solved, and the Fisher--Rao kernel leads on it (Section~\ref{sec:extended}f).

\subsection{Mixed-structure stress test (pre-specified follow-up; negative result for adaptivity)}\label{sec:stress}

If geometry selection is valuable, the adaptive mixture should dominate every pure geometry on data whose structure \emph{alternates within each sequence}. This prediction --- (a) the mixture beats all three pure geometries on the combined metric, and (b) its gates swing within each sequence, more hyperbolic in hierarchical blocks and more Euclidean in flat ones --- was written into the header of the experiment script (\texttt{exp\_mixta2.py}) before it was run; we report the outcome regardless of direction. We constructed such a task: sequences interleave marker-delimited Dyck-2 blocks (hierarchical) and copy blocks (flat), training on one block pair (lengths 22--58) and extrapolating to two pairs (76--116). The combined metric averages bracket-closing accuracy (Dyck regions) and copy accuracy (predictable copy positions); 4 arms $\times$ 5 paired seeds (Table~\ref{tab:stress}).

\begin{table}[h]
\centering\small
\begin{tabular}{lcccc}
\toprule
Arm & IN combined & OUT combined & OUT Dyck & OUT copy \\
\midrule
Euclidean & 68.2\% $\pm$ 0.4 & 53.7\% $\pm$ 0.4 & 75.3\% & 32.1\% \\
Spherical & 68.3\% $\pm$ 0.5 & 54.1\% $\pm$ 0.4 & 76.1\% & 32.2\% \\
\textbf{Hyperbolic isotropic} & \textbf{69.2\% $\pm$ 0.2} & \textbf{55.4\% $\pm$ 0.4} & \textbf{78.3\%} & \textbf{32.4\%} \\
Mixed adaptive & 68.7\% $\pm$ 0.3 & 54.5\% $\pm$ 0.4 & 76.6\% & 32.3\% \\
\bottomrule
\end{tabular}
\caption{Mixed-structure stress test.}
\label{tab:stress}
\end{table}

\textbf{The prediction failed.} The adaptive mixture scored above the Euclidean ($+0.75$ pp, $p = 0.056$, not significant) and spherical ($+0.31$ pp, $p = 0.039$, uncorrected) kernels but lost --- significantly --- to the pure hyperbolic kernel (OUT combined $\Delta = -0.91$ pp, paired $t = -7.98$, $p = 0.0013$). Moreover, the within-sequence gate swing predicted by the between-task result did not materialize: gates inside Dyck blocks (euc/sph/hyp = 0.56/0.23/0.22) barely differ from gates inside copy blocks (0.49/0.24/0.27), plausibly because the cumulative-mean gate integrates too slowly to track block boundaries.

\textbf{What the negative result suggests --- an interpretation, not a demonstrated claim.} The pure hyperbolic kernel led on the hierarchical component of the mixed task by a wide margin and did not differ on the flat one. This admits a clean geometric reading: curvature effects vanish at scales small relative to the curvature radius $1/\sqrt{|K|}$, so a hyperbolic kernel can emulate flat attention locally (by keeping distances small in units of $\sigma$), while dot-product or squared-L2 attention on linearly projected features cannot reproduce the logarithmic growth of the hyperbolic distance. Under this reading, negative curvature would function not as a hierarchy \emph{specialist} but as a \emph{superset geometry} containing approximately-flat behavior as a limiting regime, making explicit geometry mixing unnecessary at the kernel level. We flag the evidential limits of this interpretation plainly: it rests on a single mixed-structure task where the hyperbolic margin over the adaptive mixture is under one point ($-0.91$ pp, $p=0.0013$), the mixture's loss is partly attributable to a design weakness of the gate (its cumulative-mean input integrates too slowly --- a fixable engineering issue, not a geometric necessity), and the broader record of this project shows several settings where hyperbolic Fisher--Rao geometry does \emph{not} help (Sections \ref{sec:incumbents}, \ref{sec:limitations}). A convincing establishment of the superset reading would require robustness across a family of tasks with systematically varied structure. What the present data do support is narrower: on these tasks, the explicit gate added no accuracy over the pure hyperbolic kernel, while remaining valuable as a \emph{diagnostic} of which geometry the data calls for (Section~\ref{sec:gate}, between-task result). A direct confirmatory experiment --- measuring whether the model operates at smaller distance-to-$\sigma$ ratios inside flat blocks than hierarchical ones --- is left to future work.

\subsection{Comparison against established hyperbolic methods (negative result)}\label{sec:incumbents}

The experiments above compare against a Euclidean baseline. But in the domains where hyperbolic geometry is used in practice --- ontologies, taxonomies, knowledge graphs --- the incumbent is not Euclidean attention; it is the family of hyperbolic embedding methods established since 2017 \cite{nickel2017} (Poincar\'e, Lorentz). A fair claim of advantage must therefore measure the Fisher--Rao kernel against \emph{those}, on their own benchmark. We do so on the canonical task: reconstruction of the transitive closure of the WordNet mammals sub-hierarchy (1170 nodes, 6448 edges), with the standard mean-average-precision (MAP) metric, matched per-node dimensionality, and each method given its own tuned learning rate (our reimplementations reproduce the published Poincar\'e MAP $\approx 0.93$). Table~\ref{tab:wordnet} gives the results.

\begin{table}[h]
\centering\small
\begin{tabular}{lcc}
\toprule
Method & $D=5$ MAP & $D=10$ MAP \\
\midrule
Euclidean & 0.231 $\pm$ 0.001 & 0.397 $\pm$ 0.001 \\
\textbf{Poincar\'e \cite{nickel2017}} & \textbf{0.928 $\pm$ 0.005} & \textbf{0.941 $\pm$ 0.003} \\
Lorentz \cite{nickel2018lorentz} & 0.006 (diverged) & 0.764 $\pm$ 0.379$^{\dagger}$ \\
Fisher--Rao isotropic (ours) & 0.524 $\pm$ 0.028 & 0.883 $\pm$ 0.007 \\
\bottomrule
\end{tabular}
\caption{WordNet mammals reconstruction (MAP, 5 seeds). $^{\dagger}$1/5 seeds diverged; 0.953 $\pm$ 0.004 over the 4 converged.}
\label{tab:wordnet}
\end{table}

\textbf{The Fisher--Rao kernel does not beat the incumbents at pure embedding.} It dominates Euclidean by a large margin ($+0.49$ MAP at $D=10$) and is competitive at higher dimension, but it is significantly below Poincar\'e at both dimensions ($D=10$: $\Delta = -0.059$, paired $t = -12.5$, $p = 0.0002$) and collapses at low dimension where Poincar\'e remains strong. Because the isotropic Fisher--Rao parameterization with $D$ coordinates is itself a model of $D$-dimensional hyperbolic space, and MAP depends only on the ordering of distances, the gap is one of parameterization and optimization rather than of geometry: for the \emph{static embedding} problem, the Poincar\'e model space is easier to optimize than a distributional parameterization carrying a scale coordinate. Our contribution is therefore \emph{not} a better hierarchy embedder; it is a drop-in attention kernel whose value, where it exists, is in the \emph{sequence} setting (Section~\ref{sec:experiments}), not in competition with dedicated graph-embedding methods. This bounds the practical reading of the work: on the tasks those methods already serve, they remain the state of the art.

\subsection{Head-to-head against the curvature-adaptive incumbent on sequences}\label{sec:cat}

The stress test of Section~\ref{sec:stress} pits the pure kernel against \emph{our} gated mixture. The stronger test is against the incumbent of that idea: CAT \cite{lin2025cat}, which routes each token among Euclidean, hyperbolic and spherical attention branches through a differentiable per-token gate, and which has been evaluated only on graph data (knowledge-graph completion on FB15k-237 and WN18RR, plus node classification on Cora in an appendix). CAT's code is not public, so we reimplemented its central mechanism --- per-token routing among the three model-space geometries --- at our scale and label the arm ``CAT-style'': three branches in \emph{model spaces} (scaled dot product; geodesic distance in a Poincar\'e ball reached by the exponential map at the origin, with learnable curvature $c$; angular distance on the unit hypersphere), a per-token softmax gate, per-branch per-head learnable temperatures, and query/key/value/output projections shared across branches so that parameters match the baselines (414k vs.\ 413k Euclidean, 415k Fisher--Rao). Our version differs from the published design in several respects: it mixes the three branch \emph{scores} before a single softmax, whereas CAT mixes the \emph{outputs} of separate branch attentions, each followed by its own feed-forward layer (the hyperbolic and spherical branches after their own manifold aggregation and logarithmic map); its branch scores are negative squared query--key distances (geodesic in the ball, angular on the sphere) with learnable temperatures, whereas CAT scores the hyperbolic branch by $\exp(-d)$ between projected queries and the spherical branch by an exponentiated inner product; its curvature $c$ is learnable, which CAT does not describe; and it omits CAT's entropy regularization of the gate. To avoid a strawman, CAT-style alone received a pilot tuning stage on seed 1 ($lr \in \{5\times10^{-4}, 10^{-3}\}$ $\times$ $c_{\mathrm{init}} \in \{0.5, 1.0\}$; OUT accuracy 86.2--87.8\%, best $lr = 10^{-3}$, $c_{\mathrm{init}} = 1.0$), while the Euclidean and Fisher--Rao arms kept the published configuration of Section~\ref{sec:protocol} untouched. Seed 1 is shared between the pilot and the confirmatory set, which can only favor the CAT-style arm; excluding it leaves both conclusions below unchanged. The primary endpoint, written before execution, was Fisher--Rao vs.\ CAT-style on Dyck-2 OUT accuracy over five paired seeds (Table~\ref{tab:cat}).

\begin{table}[h]
\centering\small
\begin{tabular}{lccc}
\toprule
Arm (Dyck-2, 5 paired seeds) & OUT accuracy & Params & Wall-clock per run \\
\midrule
Euclidean & 85.80\% $\pm$ 0.65 & 413k & $\approx$25 s \\
CAT-style (curvature-adaptive routing, tuned) & 88.13\% $\pm$ 0.34 & 414k & $\approx$50 s \\
\textbf{Fisher--Rao isotropic (ours)} & \textbf{88.43\% $\pm$ 0.77} & 415k & $\approx$35 s \\
\bottomrule
\end{tabular}
\caption{Head-to-head against a reimplementation of the curvature-adaptive incumbent (CAT).}
\label{tab:cat}
\end{table}

\textbf{The primary contrast shows no significant difference} (Fisher--Rao $-$ CAT-style $= +0.30$ pp, paired $t = 0.69$, $p = 0.53$; 95\% CI $[-0.91, +1.51]$ pp, with no equivalence margin pre-specified; without seed 1, $+0.03$ pp, $p = 0.96$). Two further observations matter. First, CAT-style clearly beats Euclidean attention on Dyck-2 ($+2.34$ pp beyond the training length, $t = 5.07$, $p = 0.0071$; without seed 1, $+2.72$ pp, $p = 0.0038$) --- to our knowledge the first evidence that CAT's per-token routing helps on a sequence task, beyond the graph benchmarks on which it was introduced, and a reminder that the comparison is against a strong mechanism, not a weak one; the sanity contrast Fisher--Rao vs.\ Euclidean reproduces Table~\ref{tab:contrasts} exactly ($+2.64$ pp, $p = 0.0009$). Second, CAT-style shows less than half the seed-to-seed spread of the Fisher--Rao kernel ($\pm 0.34$ vs.\ $\pm 0.77$ pp): routing does not raise mean accuracy here, but it may stabilize training --- a possibility the present five seeds can raise but not establish. What the absence of a difference licenses is a parsimony argument, not a superiority claim: a single kernel of fixed curvature sign, derived rather than routed, does not differ significantly from the incumbent in mean accuracy (equivalence is not established at $n = 5$) at roughly 70\% of its wall-clock and with no gating machinery, extending the Section~\ref{sec:stress} finding from our own mixture to the incumbent itself. The caveats are the usual ones for a reimplementation --- the differences from the published design noted above could change the comparison --- and the comparison covers one task family with five seeds.

\section{Limitations}\label{sec:limitations}

\begin{itemize}
\item \textbf{Scale and scope.} CPU training throughout. The headline contrast rests on 20 paired seeds and is larger in one 19M-parameter configuration (three seeds; not a scaling trend, Section~\ref{sec:extended}a--b), and the stabilized kernel degrades less than dot-product attention beyond the training length on real Python source ($n=20$, Section~\ref{sec:extended}d) and on technical English prose ($n=10$, Section~\ref{sec:extended}e), which addresses the small-sample concern for the 20-seed, code and prose results but not the small-model one. The remaining scope limits are real: the natural-language evidence is a CPU-scale, character-level pilot on technical prose (narrative, conversational and web text, token-level modeling, and model scale are untested); in-distribution natural-data modeling consistently favors Euclidean attention; each stabilized-kernel result has one confirmation per data family (not yet an independent replication); secondary results (Sections \ref{sec:wasserstein}--\ref{sec:efficiency}, \ref{sec:gate}) still rest on 3--5 seeds; and a third pre-registered natural-data test, on Spanish clinical text, did not show the pattern (Section~\ref{sec:extended}g), so the smaller degradation beyond the training length is observed in two of three natural-data families, in small under-trained models, and is not a property of natural text in general.
\item \textbf{Baselines and training budget.} The flat baseline is dot-product attention with a fixed $1/\sqrt{d}$ temperature, whereas the Fisher--Rao kernel learns $\beta$ and per-token scales; all arms share one untuned learning rate (only the CAT-style arm was tuned) and a short budget (5 epochs on Dyck-2, at which dot-product attention reaches 95.6\% within the training length). We did not run Euclidean-distance attention \cite{kim2021lipschitz}, dot-product attention with a learned temperature, a model-space hyperbolic distance head (Lorentz or Poincar\'e \cite{gulcehre2019,chen2022fully}), or position schemes designed for extrapolation such as ALiBi, NoPE or KERPLE \cite{press2022alibi,kazemnejad2023nope,chi2022kerple}. Since the isotropic kernel is hyperbolic distance attention in half-space coordinates (Section~\ref{sec:related}), the hyperbolic-head comparison would test the parameterization rather than the geometry.
\item \textbf{Curvature and score growth are not separated.} The KL control isolates the growth of the score, not curvature, and the variant without per-token scales (a radial, logarithmically growing function of the Euclidean distance between means) keeps most of the gain in an unpaired comparison (Sections~\ref{sec:wasserstein} and~\ref{sec:sigma}). A flat kernel with logarithmic growth, such as KERPLE's logarithmic bias \cite{chi2022kerple}, and a paired fixed-scale arm are the controls that would separate them.
\item \textbf{Natural-data models are small and under-trained.} The natural-data tests use learned absolute positions up to 512 with training at 128, so rows 128--511 are never trained and the extrapolative zone measures robustness to untrained positional vectors; each model sees 4,000 windows of 128 characters for 3 epochs. On English prose both arms reach 3.60--3.64 bpc in distribution, near a smoothed character bigram (3.74) and far from a 4-gram (2.42). These results show that one under-trained model degrades less than another beyond the training length; they are not evidence about natural-language length extrapolation at scale.
\item \textbf{The depth-scaling story is refuted, not open.} A pre-registered test (Section~\ref{sec:extended}c) found the advantage does not increase with hierarchy depth --- the intuitive ``curvature matters more for deeper nesting'' narrative should not be attributed to this work.
\item \textbf{Multiple comparisons.} Only the isotropic- and anisotropic-vs-Euclidean contrasts survive strict Bonferroni correction (spherical vs.\ Euclidean, $p = 0.043$, does not); intermediate contrasts are suggestive, supported by the consistent ranking across seeds.
\item \textbf{Absolute positional embeddings.} All results except Section~\ref{sec:extended}f use learned absolute positional embeddings, whose rows beyond the training length are never trained. With relative (rotary) positions the Dyck-2 and copy advantages persist (Section~\ref{sec:extended}f), but the natural-data experiments (Section~\ref{sec:extended}d--e) have not been re-run with relative positions.
\item \textbf{Gate mechanism.} The causal cumulative-mean gate selects geometry between tasks but is too sluggish to track structure changes within a sequence (Section~\ref{sec:stress}); a local-window or learned-decay gate is the natural refinement, though the Section~\ref{sec:stress} result suggests kernel-level mixing may be unnecessary altogether.
\item \textbf{Incumbent comparison by reimplementation.} CAT has no public code; Section~\ref{sec:cat} compares against a small-scale reimplementation of its central mechanism, tuned to avoid a strawman, on one task family with five seeds. Our version departs from the published design in the respects listed in Section~\ref{sec:cat} (it mixes branch scores rather than branch outputs, scores the branches by squared query--key distances with learnable temperatures, makes the curvature learnable and omits the gate's entropy regularization); those differences could change the comparison. Nor can the reimplementation reproduce published numbers, since CAT reports none on sequences.
\item \textbf{What $\sigma$ learns is shown on synthetic hierarchies only.} The depth code and its query/key asymmetry (Section~\ref{sec:sigma}) come from two-layer models on bracket languages with five seeds per test; on program source the code did not reach the pre-registered threshold, the swap test missed significance narrowly ($p = 0.051$) and the reachability endpoint failed (12/20). The 20 head--seed cases are not independent, the untrained model is a weak null for decodability, and the intervention result would not survive a correction across the nine endpoints.
\item \textbf{Pre-registration seals are self-timestamped, and the earlier protocols carry none.} The protocols of Sections~\ref{sec:extended}f, \ref{sec:extended}g and~\ref{sec:sigma} and of Appendix~\ref{app:followup} were hashed (SHA-256) before the runs they govern (for the learned-scale tests, after a smoke test that each seal records: tensor shapes, numerical validity and, in some cases, checks of the analysis code) and the hashes are kept with them, but they were not deposited with an external registry: a matching hash shows that a protocol was not edited after hashing, not independently that the hash predates the run. The seal of Appendix~\ref{app:followup} holds the hash alone, without a timestamp; its date is the one written in the protocol and in the launcher log. For Section~\ref{sec:extended}g, the published protocol and script differ from the hashed files as described in the Reproducibility Statement, so their hashes cannot be checked against the public files. The earlier pre-registrations were written before execution but not hashed: the depth test and the code tests (Section~\ref{sec:extended}c--d) as header comments of their scripts, and the prose test, the CAT comparison and the threshold probe (Sections~\ref{sec:extended}e, \ref{sec:cat} and~\ref{sec:efficiency}) in one protocol file. The protocols are written in Spanish.
\item \textbf{Attention-only intervention.} Value aggregation and residual stream remain Euclidean; the architecture is a hybrid, not a fully Riemannian network.
\item \textbf{Not a hierarchy-embedding method.} On static graph embedding the kernel is beaten by dedicated hyperbolic methods (Section~\ref{sec:incumbents}); its advantage is specific to the sequence-attention setting and does not extend to the ontology/knowledge-graph tasks where Poincar\'e/Lorentz embeddings are already deployed.
\item \textbf{Efficiency is threshold-gated, not universal.} The parameter/compute efficiency result (Section~\ref{sec:efficiency}) only holds for $d_h \ge 16$ in the tested architecture; at smaller head widths the hyperbolic kernel is significantly \emph{less} accurate than Euclidean attention at double the dimension (at equal dimension the two do not differ significantly). The location of this threshold in other architectures, tasks, or scales is untested, and the mechanism behind it (why the crossover occurs specifically around $d_h \approx 16$) is not explained by the present experiments --- a natural target for follow-up work rather than a settled account.
\item \textbf{Other efficiency/advantage hypotheses tested and refuted.} Beyond the graph-embedding result above, two further follow-up investigations (outside the scope of the attention mechanism itself, using a distance-based classifier rather than attention) found no advantage for the Fisher--Rao geometry: it was \emph{not} a usable predictive-uncertainty signal via the scale parameter $\sigma$ (AUC-ROC $\approx 0.49$, chance level), and in few-shot hierarchical classification it was \emph{outperformed} by both Euclidean and Poincar\'e baselines at low sample counts ($k \le 10$/class; $p = 0.0003$ vs.\ Poincar\'e at $k = 10$) --- the opposite of the pattern one might expect from hyperbolic geometry's known data efficiency in the embedding literature. These are reported for completeness and to bound enthusiasm about the generality of Section~\ref{sec:efficiency}'s efficiency result: it is specific to the attention mechanism and the tested regime, not a general property of the geometry.
\end{itemize}

\section{Related Work and Positioning of Contributions}\label{sec:related}

This work sits within an active and crowded literature; we state carefully what is prior and what, to our knowledge, is not.

\textbf{Adaptive/mixed-curvature attention is prior work, not a contribution of this paper.} The idea of learning, per token, to route among Euclidean, hyperbolic, and spherical attention branches via a differentiable gate is introduced by CAT \cite{lin2025cat} and related to earlier mixed-curvature transformers \cite{cho2023curve} and mixture-of-curvature-experts LLMs \cite{he2025helm}. Our gated mechanism (Sections \ref{sec:mixed}, \ref{sec:gate}) is a re-instantiation of this idea, not a novel one; we use it as a \emph{diagnostic probe} of which geometry a task calls for, and our Section~\ref{sec:stress} stress test contributes a (negative) empirical finding about its necessity rather than the mechanism itself. Section~\ref{sec:cat} adds a direct comparison against a reimplementation of CAT on Dyck-2, where CAT beats Euclidean attention and does not differ significantly from the pure Fisher--Rao kernel.

\textbf{Gaussian/distributional tokens with distance-aware attention are also prior work.} Representing tokens as Gaussians and scoring by a distribution distance appears in sequential recommendation \cite{fan2022stosa} and in stochastic vision transformers \cite{erick2023stochastic}, both using the Wasserstein distance, echoing probabilistic word embeddings \cite{vilnis2015}; density-adaptive attention \cite{ioannides2024daam} uses Gaussians differently, modulating features with a learnable Gaussian density (learnable mean offset and scale) rather than scoring a distance between token distributions. Our point of departure from \cite{erick2023stochastic} is the \emph{choice of distribution distance}: Wasserstein--Bures yields a non-hyperbolic optimal-transport geometry, whereas the Fisher--Rao geodesic yields genuine negative curvature. Section~\ref{sec:wasserstein} shows this is not a cosmetic distinction --- with token representation held fixed, the Fisher--Rao kernel outperforms Wasserstein by $+1.52$ pp ($p=0.029$), a larger effect than that of using Gaussian tokens in the first place.

\textbf{What we believe is new --- as a synthesis, not an invention.} Relative to the above, the specific elements we could not locate in prior work are: (i) deriving the attention geometry from Chentsov invariance across \emph{distributional families} --- categorical $\to$ spherical, isotropic Gaussian $\to$ hyperbolic --- as members of one Fisher--Rao formalism rather than as independent model-space choices (CAT and \cite{cho2023curve,he2025helm} use Poincar\'e/Lorentz/hypersphere model spaces, not the Fisher--Rao metric on distributions, and do not invoke Chentsov); (ii) the \emph{exact} isotropic-Gaussian Fisher--Rao geodesic (Eq.~\ref{eq:dist}) as an attention score. Under the standard isometry from the upper half-space to the hyperboloid, $P(\mu,\sigma) = \big((\sigma^2 + \lVert\mu\rVert^2/2 + 1)/(2\sigma),\ \mu/(\sqrt{2}\,\sigma),\ (\sigma^2 + \lVert\mu\rVert^2/2 - 1)/(2\sigma)\big)$, one has $\cosh(\dFR/\sqrt{2}) = -\langle P_i, P_j\rangle_L$. The kernel is therefore hyperbolic geodesic-distance attention in half-space coordinates, and its one-matmul cost is that of Lorentz-model attention \cite{gulcehre2019,nickel2018lorentz,chen2022fully}. What is specific here is the chart --- tokens parameterized by a linear mean head and a softplus scale head, with the scale read as a Gaussian standard deviation and the known closed-form distance \cite{costa2015} --- not the distance computation; (iii) evaluation beyond the training length --- formal languages, program source and English prose --- rather than in-distribution benchmarks (knowledge graphs, vision, translation); and (iv) an analysis of what the learned scales encode inside the kernel, including an opposite-direction query/key depth code (Section~\ref{sec:sigma}). We stress that this is a novel \emph{combination and analysis} of known ingredients.

\textbf{Prior-art search conducted.} A multi-source search (arXiv API, OpenReview API, Semantic Scholar, a manual Google Scholar sweep, plus general web search; July--August 2026, with the arXiv and OpenReview exact-phrase sweeps repeated in October 2026 and an arXiv title search in November 2026; query log and verdicts on the nearest hits in \texttt{verificacion\_claude/}\allowbreak\texttt{busqueda\_novedad.md}) found no work computing attention scores as Fisher--Rao geodesic distances between distributional tokens. The exact phrases ``Fisher--Rao geodesic attention'', ``information geometric attention'', and ``Gaussian hyperbolic attention'' return no relevant results on arXiv, OpenReview, and Google Scholar; ``Fisher--Rao attention'' returns exactly one Google Scholar hit --- concurrent work using the Fisher--Rao distance to \emph{measure} attention shifts in a standard multimodal transformer, assign functional roles to tokens and reweight reinforcement-learning advantages accordingly \cite{lin2026dyco} --- a training signal derived from standard attention, not an attention kernel. The nearest neighbors beyond those already discussed are: a concurrent \emph{theoretical} analysis of the Fisher--Rao geometry of attention outputs, proving that ungated attention is confined to flat statistical manifolds while multiplicative gating enables curvature \cite{bathula2026gating} --- theory about the geometry \emph{of} standard attention, where the present work builds attention \emph{from} geodesic distances, and a result that dialogues directly with our Section~\ref{sec:stress} observation that, on the tasks tested, an explicit gate added no accuracy over the pure hyperbolic kernel; a related analysis tying rotary-position-embedding temperature to the Fisher--Rao geometry of attention distributions \cite{mtetwa2026rope}; a place-recognition pipeline using Fisher--Rao for relation features alongside hyperbolic self-attention \cite{shang2026riemannian}; and a wireless-networks model using a Fisher--Rao-motivated hypersphere projection \emph{before} standard attention \cite{jiu2026rieif}, adjacent to our spherical case (Section~\ref{sec:spherical}). The November title search added five neighbors, none of which scores attention by a Fisher--Rao distance between distributional tokens: a theoretical result that a standard softmax attention head can compute the Bayes posterior for exponential-family observations, with the log-likelihood written as a negative Bregman divergence \cite{rubacher2026attention} --- the dually flat geometry of our KL control (Section~\ref{sec:wasserstein}), not the Fisher--Rao geodesic; exponential family attention \cite{wibisono2025exponential}, in which the family is the likelihood of each observation and its natural parameter is read off a standard dot-product attention layer; QUEST \cite{govindarajan2026quest}, which normalizes keys onto a hypersphere and leaves queries unnormalized, a neighbor of our spherical member (Section~\ref{sec:spherical}) with a dot-product rather than a geodesic score; and two concurrent theoretical preprints without experiments, one scoring attention by a pair-dependent Mahalanobis distance under learned per-token metrics \cite{song2026riemannian}, the other replacing the softmax by an inverse-distance weight with the Euclidean, Poincar\'e-ball or great-circle distance between points of model spaces chosen \emph{a priori} \cite{ge2026georesolver}. Hyperbolic neural networks define layers, including attention, in model spaces \cite{ganea2018hnn,shimizu2021hnnpp,chen2022fully}, and Hypformer builds a full transformer in the Lorentz model with a linear-time self-attention \cite{yang2024hypformer}. Hyperbolic distance itself has been used as an attention score before, outside the distributional setting, for example: hyperbolic attention networks weight query--key pairs by a decreasing function of their distance on the hyperboloid, $f(-\beta\, d_{\mathbb{H}} - c)$ with $f$ exponential (normalized as a softmax) or sigmoid \cite{gulcehre2019}; a hyperbolic vision transformer scores by a negative squared Poincar\'e-ball distance \cite{feinashley2024hvt}; and, in concurrent and independent work identified by an updated search in October 2026, a fully hyperbolic vision transformer scores by the negative squared Lorentzian geodesic distance, in a fused kernel that never stores the score matrix \cite{bdeir2026polar}. The first has, in its softmax variant, the functional form of our Laplacian score (Section~\ref{sec:ablations}), and the other two that of our squared-distance score (Section~\ref{sec:iso}); in all three the form is reached from the opposite direction: there, tokens are points of a hyperbolic model space chosen \emph{a priori}; here, tokens are Gaussian distributions and the hyperbolic geometry follows from the Fisher--Rao metric, fixed up to scale by Chentsov's theorem once the family is chosen, the same construction yielding the spherical member of the family (Section~\ref{sec:spherical}). Residual caveats: the Semantic Scholar sweep was partial (rate limits), the Google Scholar sweep dates from August 2026, coverage is English-only, and \cite{rubacher2026attention} was read from its abstract only.

Curvature-based analysis of information flow in networks \cite{topping2022} is related in spirit but addresses a different problem (graph rewiring against over-squashing); the present mechanism uses fixed constant-curvature manifolds with learned movement of representations, not dynamic metric evolution. In the source-code domain specifically, concurrent work introduces a hyperbolic GNN over abstract syntax trees with a Lorentz-model geodesic attention module for graph message-passing \cite{zhou2026hypecode} --- further evidence for hyperbolic geometry's fit to code structure, and a natural point of comparison for our sequence-level kernel on code, where it is worse in distribution but degrades less beyond the training length once norm-capped (Section~\ref{sec:extended}d).

\section{Conclusion}

Equipping attention with exact Fisher--Rao geodesic kernels makes the geometry of the latent space a principled and cheap degree of freedom. On Dyck-2, the negatively curved kernel learns the nested structure best, within and beyond the training length: the parameter-matched hyperbolic-isotropic kernel --- at $\approx 1.4\times$ the end-to-end training time of standard attention --- provides the strongest gains, and ablations attribute the effect to the geodesic kernel --- its slow growth and, to an extent not yet separated, its curvature --- rather than to the distributional representation or to the local information geometry it shares with a KL kernel (Section~\ref{sec:wasserstein}). A positive-definite Laplacian form of the same geodesic kernel retains most of the advantage (Section~\ref{sec:ablations}), keeping open a route to linear-time variants through random features for hyperbolic kernels \cite{yu2023hyla}. Above a head-width threshold, this translates into a genuine resource-efficiency result: accuracy comparable to Euclidean attention at roughly a quarter of the parameters and, at the largest scale tested, less wall-clock compute (Section~\ref{sec:efficiency}) --- comparable performance from a smaller model, not better performance. Within that regime the data are consistent with reading negative curvature as a safe default whose small-scale limit approximates flat behavior (Section~\ref{sec:stress}), under which an explicit geometry gate proves diagnostic rather than necessary --- a reading that survives a head-to-head with the curvature-adaptive incumbent itself (no significant difference, Section~\ref{sec:cat}), though establishing it beyond the present tasks remains open.

The positive results are: the hyperbolic-vs-dot-product advantage on Dyck-2, robust over 20 seeds ($+2.10$ pp, 95\% CI $[1.74, 2.46]$) and with relative (rotary) positions in a separately pre-registered 20-seed test (Section~\ref{sec:extended}f); a larger advantage in one $46\times$ (19M-parameter) configuration; threshold-gated parameter efficiency; ablations locating the gain in the geodesic kernel rather than in the distributional representation or the local information metric; smaller degradation beyond the training length on program source ($n{=}20$, $p = 0.0037$) and English technical prose ($n{=}10$, $p = 0.0002$) after a diagnosis-driven stabilization; and a depth code in the learned scales (Section~\ref{sec:sigma}). The pre-specified hypotheses that failed include: geometry gating (Section~\ref{sec:stress}), $\sigma$ as an uncertainty estimator (Section~\ref{sec:limitations}), static hierarchy embedding against Poincar\'e/Lorentz (Section~\ref{sec:incumbents}), few-shot classification (Section~\ref{sec:limitations}), depth-amplification of the advantage (Section~\ref{sec:extended}c), the original kernel's far-extrapolation endpoint on code (twice, Section~\ref{sec:extended}d, and again on a second code corpus, Appendix~\ref{app:followup}), the crossed pattern on Spanish clinical text (Section~\ref{sec:extended}g), and three endpoints about $\sigma$ (the swap, the code on program source and the reachability account; Section~\ref{sec:sigma}). In-distribution natural-data modeling favors dot-product attention throughout, as predicted for prose.

We are deliberate about scope. The practical envelope these experiments support is specific: \emph{sequence models learning structured data, including beyond their training length, where parameter or memory budgets matter} --- with in-distribution tasks, static embedding, and hardware-fused attention pipelines remaining outside it. Effects are single-digit percentage points or hundredths of bits per character; the efficiency gain reverses below $d_h \approx 16$ for reasons still unexplained; the natural-data results of the norm-capped kernel have one adequately powered confirmation per data family and await replication at scale on general-domain text. Apart from performance, the isotropic Gaussian family gives each token an explicit scale that point tokens lack, and what that scale learns can be asked: on bracket languages it matches the geometric reading of $\sigma$ as a height only in part, with second-layer query scales narrowing with nesting depth, as that reading suggests, and key scales widening (Section~\ref{sec:sigma}). In sum, Fisher--Rao geodesic attention is a principled and inexpensive way to give a Transformer's sequence attention a geometry derived from a statistical model --- resource-efficient above a dimension threshold --- with small pre-registered effects on two kinds of natural data, at a scale far below practical language models, and a documented map of where it does not help. The latent space need not be flat: its geometry can be derived from a statistical model, stabilized, and tested to its boundaries, which is what this work has tried to do.

\section*{Use of AI Tools}

The original idea and the direction of the research are the authors'. AI assistants were used throughout, under the authors' direction. Claude (Anthropic) was used to draft protocols and to write and review the experiment code; as support to run the experiments on the authors' machines and, as a third reviewer alongside the two authors, to check them; as a second reviewer in the literature search; and to prepare the manuscript text from the authors' own drafts and proposals. The second author also used DeepSeek and Qwen to review text, to run experiments and for cross-checks. The authors reviewed and approved all content and take full responsibility for it.

\section*{Reproducibility Statement}

All attention experiments except the tests of Section~\ref{sec:sigma} (scripts available from the authors) and of Section~\ref{sec:extended}g (corpus not redistributed) can be re-run on a laptop CPU ($\approx$5 h for the v1.1 set; a re-execution of the 14 experiments of the v1.1 README by the second author, on separate hardware, is reported in Appendix~\ref{app:repro}) from the scripts in the public repository (\url{https://github.com/quinelen/fisher-rao-attention}; archived at the concept DOI \href{https://doi.org/10.5281/zenodo.23424909}{10.5281/zenodo.23424909}): the main 3-geometry benchmark, the approximation arm, the isotropic kernel with speed micro-benchmark and ablations, the mixed-curvature gating experiments, the mixed-structure stress test, the geometry-vs-representation ablation, the parameter/compute efficiency sweep, the WordNet comparison against Poincar\'e/Lorentz (our reimplementations reproduce the published Poincar\'e MAP $\approx 0.93$), the 20-seed extension and 19M-parameter scale probe, the three pre-registered length-extrapolation tests on real Python source with the failure-mode diagnosis leading to the stabilized kernel (the code corpus is read from the standard library of the running interpreter --- all code-corpus runs reported in the main text used CPython 3.9.6, and Appendix~\ref{app:repro} reports runs on other interpreters --- and a manifest lists its files and SHA-256 hashes), the pre-registered depth-stratified Dyck-4 test, the cap-robustness analysis, the threshold-mechanism probe, the pre-registered English-prose extrapolation test (its corpus is rendered deterministically from the local system manual, so its exact content depends on the installed pages; a manifest lists the 274 pages, the operating-system version and SHA-256 hashes against which a re-rendering can be checked), the CAT-style reimplementation with its pilot tuning and head-to-head, the KL/Bregman and geodesic-Laplacian kernel ablations, and the pre-registered relative-position test, whose protocol was sealed with SHA-256 hashes before execution --- with raw outputs in the corresponding result files. The pre-registrations of the English-prose test, the CAT comparison and the threshold-mechanism probe are in \texttt{verificacion\_claude/}\allowbreak\texttt{PREREGISTRO\_FASE2\_CPU.md}; the relative-position protocol is in \texttt{verificacion\_claude/}\allowbreak\texttt{PREREGISTRO\_ROPE.md}, with its pre-execution SHA-256 seal in \texttt{verificacion\_claude/}\allowbreak\texttt{SELLO\_PREREGISTRO\_ROPE.txt}. The pre-registered tests of what $\sigma$ learns (Section~\ref{sec:sigma}) are in \texttt{verificacion\_claude/}\allowbreak\texttt{sigma/}, with protocols, pre-execution SHA-256 seals, logs and results; their scripts, whose hashes are in the seals, are available from the authors. The Spanish clinical-text test (Section~\ref{sec:extended}g) is in \texttt{verificacion\_claude/}\allowbreak\texttt{guias\_es/}, with its protocol, seal, script, a manifest listing the 82 guidelines and the SHA-256 hashes of the training and test texts, and the list of their public URLs; the guideline texts are public, but neither the database they were extracted from nor the corpus text is redistributed. The published script of Section~\ref{sec:extended}g differs from the sealed one in two lines (local paths), and the published protocol omits two administrative sentences about the database copy; the hashes of these two files therefore do not match the seal, while that of the manifest does. The seals are self-timestamped (that of Appendix~\ref{app:followup} holds the hash alone), and the pre-registrations before the relative-position test carry none (Section~\ref{sec:limitations}). Of the two refuted follow-up hypotheses referenced in Section~\ref{sec:limitations}, which used a distance-based classifier outside the attention-mechanism scope, the few-shot hierarchical classification test is included; the uncertainty-calibration test used a separate codebase and is not included. Seeds, configurations, and statistical procedures are specified in-line in each script.

\appendix
\section{Re-execution on Separate Hardware and a Pre-Registered Follow-Up}\label{app:repro}

This appendix reports a re-execution of the 14 experiments listed in the v1.1 README, carried out by the second author on a different machine from the one used for the original runs, together with a partial run in a third environment and one pre-registered follow-up test. All scripts were run unmodified from the tagged release v1.1 (commit \texttt{b7cd4b4}). Wall-clock fields are excluded from every comparison below; all other numeric outputs are compared value by value against the result files published with v1.1. The re-execution covers that list only. Not re-run are, in particular, the 20-seed extension and the 19M-parameter probe (Section~\ref{sec:extended}a--b), the Dyck-4 depth test, the cap sensitivity analysis and the threshold probe, and the experiments added in v1.2 (Sections~\ref{sec:extended}g and~\ref{sec:sigma}); the code tests with the original kernel were run only as reported in Sections~\ref{app:colab} and~\ref{app:followup}, on another interpreter.

\subsection{Environments}\label{app:env}

\begin{table}[h]
\centering\small
\resizebox{\textwidth}{!}{%
\begin{tabular}{llll}
\toprule
 & Original runs & Re-execution & Cross-platform run \\
\midrule
Hardware & laptop, Apple silicon (arm64) & Mac Studio, Apple M4 Max, 16 cores & Google Colab, x86-64 CPU \\
Operating system & macOS 26.5.2 & macOS 27.0.1 & Linux \\
Python & CPython 3.9.6 & CPython 3.9.6 (system) & CPython 3.13 \\
PyTorch & 2.8 & 2.8.0 (NumPy 1.26.4, SciPy 1.13.1) & 2.11.0 (NumPy 1.26.4, SciPy 1.16.3) \\
Scope & all experiments & the 14 experiments of the README & three scripts$^\dagger$ \\
\bottomrule
\end{tabular}%
}
\caption{Environments. $^\dagger$\texttt{fast\_gaussian\_test.py}, \texttt{exp\_codigo\_extrapolativo.py}, \texttt{exp\_rope.py}. The Colab session had a GPU attached; these scripts do not use it.}
\label{tab:app-env}
\end{table}

The re-execution ran the 14 experiments of the README block ``Reproducing the benchmark'' in that order and took 2\,h\,10\,min of wall-clock time on CPU, against the $\approx$5\,h stated in the Reproducibility Statement. Thread counts are those fixed inside each script. Other processes shared the machine during part of the run; this affects wall-clock time only, since every result below is deterministic on CPU.

\subsection{Experiments on synthetic and fixed data are bit-identical}\label{app:identical}

Every experiment whose data are generated in-line or fixed (the WordNet hierarchy) reproduced the published numbers exactly: each accuracy, loss, gate weight, test statistic and $p$-value is the same floating-point value (Table~\ref{tab:app-identical}). These numbers therefore stand as reported, including the isotropic kernel's 88.43\% on Dyck-2 (Section~\ref{sec:main}) and the pre-registered relative-position endpoint ($+1.82$ pp, $p = 6\times10^{-5}$, 16/20 seeds; Section~\ref{sec:extended}f).

\begin{table}[h]
\centering\small
\resizebox{\textwidth}{!}{%
\begin{tabular}{lcccc}
\toprule
Experiment (script) & Section & Values compared & Identical & Subset$^{\ast}$ \\
\midrule
Three geometries (\texttt{exp\_tres\_geometrias.py}) & \ref{sec:main}--\ref{sec:stats} & 48 & 48 & 45 \\
Ricci--Mahalanobis arm (\texttt{exp\_ricci\_addendum.py}) & \ref{sec:main} & 16 & 16 & 15 \\
Isotropic kernel, log-Bhattacharyya arm (\texttt{fast\_gaussian\_test.py acc}) & \ref{sec:main}, \ref{sec:ablations} & 30 & 30 & 20 \\
Mixed-curvature gating (\texttt{exp\_mixta.py}) & \ref{sec:gate} & 234 & 234 & 220 \\
Mixed-structure stress test (\texttt{exp\_mixta2.py}) & \ref{sec:stress} & 86 & 86 & 6 \\
Fisher--Rao vs.\ Wasserstein--Bures (\texttt{exp\_wasserstein\_vs\_fisherrao.py}) & \ref{sec:wasserstein} & 15 & 15 & 15 \\
WordNet vs.\ Poincar\'e/Lorentz (\texttt{exp\_wordnet\_incumbentes.py}) & \ref{sec:incumbents} & 80 & 80 & 80 \\
Dimension sweep (\texttt{exp\_eficiencia\_dimensional.py}) & \ref{sec:efficiency} & 80 & 80 & 10 \\
KL/Bregman and geodesic-Laplacian ablations (\texttt{exp\_ablacion\_kernel.py}) & \ref{sec:ablations}--\ref{sec:wasserstein} & 74 & 74 & 64 \\
CAT-style head-to-head (\texttt{exp\_cat\_incumbente.py}) & \ref{sec:cat} & 51 & 51 & 26 \\
Relative positions, Dyck-2 and copy (\texttt{exp\_rope.py}) & \ref{sec:extended}f & 262 & 262 & 178 \\
\midrule
\textbf{Total} & & \textbf{976} & \textbf{976} & \textbf{679} \\
\bottomrule
\end{tabular}%
}
\caption{Experiments on synthetic and fixed data: every numeric value of the re-executed result files against the published ones, excluding only the wall-clock fields. $^{\ast}$Values counted by the progress script used during the run, whose name filter also skipped some fields that are not timings (for example parameter counts); all of them are identical as well.}
\label{tab:app-identical}
\end{table}

\subsection{Computational cost depends on hardware for per-coordinate kernels only}\label{app:cost}

Table~\ref{tab:app-speed} repeats the micro-benchmark of Section~\ref{sec:cost} on the three machines. The overhead of the isotropic kernel is stable (2.3--2.8$\times$). The overhead of the per-coordinate formulations varies up to four-fold across machines (21--110$\times$), which is consistent with the explanation in Section~\ref{sec:cost} that their cost is dominated by materializing the $[B, H, L, L, d_h]$ tensor. The $28.5\times$ of the anisotropic kernel in Table~\ref{tab:speed} (original hardware) is thus the lowest of the three measurements, which is why the main text reports these ratios as ranges; the claim that the isotropic reduction removes the dominant cost holds on all three.

\begin{table}[h]
\centering\small
\resizebox{\textwidth}{!}{%
\begin{tabular}{lccc}
\toprule
Kernel & Original & Re-execution (M4 Max) & Colab (x86-64) \\
\midrule
Euclidean & 16.7 ($1\times$) & 2.9 ($1\times$) & 16.7 ($1\times$) \\
Hyperbolic isotropic (matmul decomposition) & 38.5 ($2.3\times$) & 8.0 ($2.8\times$) & 41.7 ($2.5\times$) \\
Log-Bhattacharyya approximation (5-D broadcast) & 348.4 ($21\times$) & 81.0 ($28\times$) & 1123.6 ($67\times$) \\
Hyperbolic anisotropic exact (5-D broadcast) & 476.1 ($28.5\times$) & 116.7 ($40\times$) & 1841.1 ($110\times$) \\
\bottomrule
\end{tabular}%
}
\caption{Speed micro-benchmark (\texttt{fast\_gaussian\_test.py bench}: one attention module, forward+backward, $B{=}32$, $L{=}60$, $d{=}128$, one CPU thread), ms per step and ratio to Euclidean attention on the same machine.}
\label{tab:app-speed}
\end{table}

\subsection{Natural-data experiments under drifted corpora}\label{app:natural}

The two natural-data corpora are built from files of the running system, and both drifted between the original runs and the re-execution.
\emph{Program source} (Section~\ref{sec:extended}d): two of the 46 standard-library files differ (\texttt{\_osx\_support.py}, \texttt{\_sysconfigdata\_\_darwin\_darwin.py}); the four held-out files are the same (37{,}771 characters), the training text grows from 1{,}195{,}931 to 1{,}199{,}078 characters and the vocabulary from 97 to 98 symbols.
\emph{Technical prose} (Section~\ref{sec:extended}e): macOS 27.0.1 installs the same 274 section-1 pages plus 5 new ones. Page text barely changed (median change in length 0.1\%), but the 5 additional pages shift the deterministic 90/10 split by page, so only 1 of the 28 held-out pages coincides with the original test set. The re-execution therefore evaluates on almost entirely different held-out pages, most of which were training pages in the original run (2.80M training and 0.35M test characters, against 2.73M and 0.42M).

Both pre-registered endpoints are met again (Table~\ref{tab:app-natural}), with the same crossed pattern: worse in distribution, better beyond the training length. The secondary endpoint at $L{=}256$ also holds in both re-executed tests ($\Delta = -0.130$, $p = 0.0003$ on code; $\Delta = -0.074$, $p = 0.0002$ on prose).

\begin{table}[h]
\centering\small
\resizebox{\textwidth}{!}{%
\begin{tabular}{llcccccc}
\toprule
Test & Run & $\Delta$ bpc & $t$ & $p$ & Seeds won & Spread & In-dist.\ $\Delta$ \\
\midrule
Program source, seeds 16--35 ($n{=}20$) & Original & $-0.084$ & $-3.30$ & 0.0037 & 15/20 & $0.97\times$ & $+0.040$ \\
 & Re-execution & $-0.076$ & $-2.63$ & 0.016 & 16/20 & $0.86\times$ & $+0.044$ \\
Technical prose, seeds 46--55 ($n{=}10$) & Original & $-0.058$ & $-6.02$ & 0.0002 & 10/10 & $1.14\times$ & $+0.032$ \\
 & Re-execution & $-0.054$ & $-4.69$ & 0.0011 & 9/10 & $0.96\times$ & $+0.029$ \\
\bottomrule
\end{tabular}%
}
\caption{Stabilized kernel vs.\ Euclidean attention, extrapolative zone at $L{=}512$. Spread: seed-to-seed standard deviation of the Fisher--Rao arm over that of the Euclidean arm (pre-specified limit $1.5\times$). In-dist.\ $\Delta$: the $L{=}128$ control.}
\label{tab:app-natural}
\end{table}

\subsection{Cross-platform run}\label{app:colab}

On Google Colab (x86-64, CPython 3.13, PyTorch 2.11), the relative-position test of Section~\ref{sec:extended}f reproduced the copy task to within 0.11 pp per seed, with the Euclidean arm identical at the reported precision ($+5.60$ pp, $p = 0.0002$, 16/20 seeds). On Dyck-2, individual seeds moved by up to 2.1 pp, but the pre-registered endpoint is met again: $+1.59$ pp (95\% CI $[0.68, 2.50]$), $p = 0.0016$, 14/20 seeds. Bit-exact agreement held in the re-execution (Appendix~\ref{app:identical}), which matched the original architecture and PyTorch version; across architectures and versions, individual seeds change while the conclusions do not.

The same environment also ran \texttt{exp\_codigo\_extrapolativo.py}, the code-extrapolation test with the original kernel and seeds 1--5, whose corpus was in that case built from the CPython 3.13 standard library. Unlike the original runs, it met the $L{=}512$ endpoint: $\Delta = -0.110$ bpc, $p = 0.012$, 5/5 seeds, spread $0.43\times$. Because this was an unplanned observation on five seeds, it was tested with the pre-registered follow-up below rather than reported as a finding.

\subsection{Pre-registered follow-up: the original kernel on a CPython 3.13 corpus}\label{app:followup}

\textbf{Hypothesis.} With a corpus built from the CPython 3.13 standard library, the original kernel (no $\mu$-norm cap) predicts better than Euclidean attention in the extrapolative zone at $L{=}512$.

\begin{sloppypar}
\textbf{Protocol.} Written and sealed with a SHA-256 hash (\texttt{96cbe2bc\ldots a494a9722}) before any training; the launcher verified the hash before running. Scripts unmodified: \texttt{exp\_codigo\_extrapolativo.py} (seeds 1--5, descriptive) and \texttt{exp\_codigo\_extrapolativo\_rep.py} (seeds 6--15, confirmatory: the same seeds as the failed replication of Section~\ref{sec:extended}d). Primary endpoint: extrapolative-zone bpc at $L{=}512$, two-sided paired $t$-test, $p < 0.05$, $n = 10$. Environment: CPython 3.13.15, PyTorch 2.14.1, NumPy 2.5.4, SciPy 1.18.1, on the re-execution machine; corpus of 1{,}099{,}928 training and 113{,}196 test characters, 111 symbols, with its manifest recorded before training.
\end{sloppypar}

\begin{table}[h]
\centering\small
\begin{tabular}{llccccc}
\toprule
Seeds & Zone & $\Delta$ bpc & $t$ & $p$ & Seeds won & Spread \\
\midrule
1--5 (descriptive) & extrapolative, $L{=}512$ & $-0.103$ & $-4.67$ & 0.0095 & 5/5 & $0.46\times$ \\
 & extrapolative, $L{=}256$ & $-0.144$ & $-5.04$ & 0.0073 & 5/5 & $0.35\times$ \\
 & in distribution, $L{=}128$ & $+0.017$ & $+3.34$ & 0.029 & 0/5 & $0.92\times$ \\
\midrule
\textbf{6--15 (primary)} & \textbf{extrapolative, $\boldsymbol{L{=}512}$} & $\mathbf{-0.023}$ & $\mathbf{-0.97}$ & \textbf{0.36} & \textbf{7/10} & $\mathbf{1.48\times}$ \\
 & extrapolative, $L{=}256$ & $-0.051$ & $-2.44$ & 0.038 & 8/10 & $0.99\times$ \\
 & in distribution, $L{=}128$ & $+0.038$ & $+5.26$ & 0.0005 & 0/10 & $1.38\times$ \\
\midrule
1--15 (post hoc) & extrapolative, $L{=}512$ & $-0.050$ & $-2.51$ & 0.025 & 12/15 & $1.32\times$ \\
\bottomrule
\end{tabular}
\caption{Pre-registered follow-up with the original kernel on a CPython 3.13.15 corpus. The primary endpoint (bold row) was not met.}
\label{tab:app-followup}
\end{table}

\textbf{The primary endpoint was not met} (Table~\ref{tab:app-followup}). Following the interpretation fixed in the protocol, the Colab observation of Appendix~\ref{app:colab} is treated as not replicated, and the account of Section~\ref{sec:extended}d stands: the original kernel does not reliably meet the far-extrapolation endpoint, the moderate-extrapolation zone ($L{=}256$) favors it, and its seed-to-seed spread rises toward the stability limit, which is the failure mode the $\mu$-norm cap was introduced to correct. The follow-up thus reproduces that pattern on a second code corpus. Seeds 1--5 were favorable to the original kernel in all three corpora (CPython 3.9.6: $\Delta = -0.077$, 4/5; Colab, CPython 3.13: $-0.110$, 5/5; this follow-up: $-0.103$, 5/5), a reminder that five seeds can mislead in this setting.

\subsection{Open item: absolute bpc level in the CPython 3.13.15 environment}\label{app:open}

Paired comparisons in the follow-up are internally valid, since both arms share the environment, but the absolute level of the bits-per-character values is higher than in the other two environments and remains unexplained. For seeds 1--5, the Euclidean arm reaches 4.40 bpc in the extrapolative zone at $L{=}512$ and 4.04 in distribution, against 3.88 and 3.43 on Colab (CPython 3.13, PyTorch 2.11) and 3.83 and 3.25 in the original runs (CPython 3.9.6, PyTorch 2.8). The candidates are the corpus, which differs between CPython 3.13 patch releases, and the PyTorch version. Separating them requires running PyTorch 2.8 on the CPython 3.13.15 corpus; we leave this as an open note.

\subsection{Artifacts}\label{app:artifacts}

The run logs, per-experiment result files, corpus manifests, the sealed protocol of Appendix~\ref{app:followup} with its hash file, and the wrapper and comparison scripts used for the re-execution are archived in the \texttt{reproduccion/} folder of the public repository and in the corresponding Zenodo version. The wrappers create the Python environment, run the unmodified scripts, record the environment and compare outputs; the wrapper of the follow-up also writes the corpus manifest and computes the paired tests of Table~\ref{tab:app-followup}. Two items are archived in a weaker form. For the Colab run, only the console output was kept; its result files were lost with the session. The manifest of the re-executed code corpus was regenerated on the same machine and interpreter after the run, which had stored only a summary; it matches that summary in file count, character counts, vocabulary and the two files that differ from the published corpus. The corpus texts are not redistributed: the manifests record character counts and SHA-256 hashes against which a regenerated corpus can be checked.

\end{document}